\documentclass{article}

\usepackage{microtype}
\usepackage{graphicx}
\usepackage{subcaption}
\usepackage{booktabs} 

\usepackage{hyperref}

\usepackage[preprint]{icml2026}

\usepackage{amsmath}
\usepackage{amssymb}
\usepackage{mathtools}
\usepackage{amsthm}

\usepackage[capitalize,noabbrev]{cleveref}

\usepackage{placeins}

\theoremstyle{plain}

\theoremstyle{definition}

\theoremstyle{remark}

\usepackage[textsize=tiny]{todonotes}

\begin{document}

\twocolumn[

\icmltitle{Debunking Grad-ECLIP: A Comprehensive Study on Its Incorrectness and Fundamental Principles for Model Interpretation}

  \icmlsetsymbol{equal}{*}

  \begin{icmlauthorlist}
    \icmlauthor{Yongjin Cui}{sch}
    \icmlauthor{Xiaohui Fan}{sch}
  \end{icmlauthorlist}

  \icmlaffiliation{sch}{Zhejiang University}

  \icmlcorrespondingauthor{Xiaohui Fan}{fanxh@zju.edu.cn}

  \icmlkeywords{Machine Learning, ICML}

  \vskip 0.3in
]

\printAffiliationsAndNotice{}  

\begin{abstract}

Grad-ECLIP is published at ICML 2024 and represents a new Transformer interpretation technical route (intermediate features-based). First, this paper demonstrates that the intermediate features-based technical route is not a novel one. Based on the existing attention-based route, we have developed Attention-ECLIP, which is completely equivalent to Grad-ECLIP but with simpler computation. Both through formal derivation and experimental validation, we prove that the intermediate feature-based route represented by Grad-ECLIP is actually an equivalent variant of the attention-based route. Next, this paper demonstrates that the Grad-ECLIP method is flawed. The model interpretation results obtained by Grad-ECLIP are not those of the original model, and the interpretation results are misaligned with the model's performance. We analyze the causes of Grad-ECLIP's flaws and propose, or rather, explicitly emphasize two fundamental principles that model interpretation should adhere to in order to avoid similar errors.

\end{abstract}

\section{Introduction}
\begin{figure*}[]
    \centering
    \includegraphics[width=1.6\columnwidth]{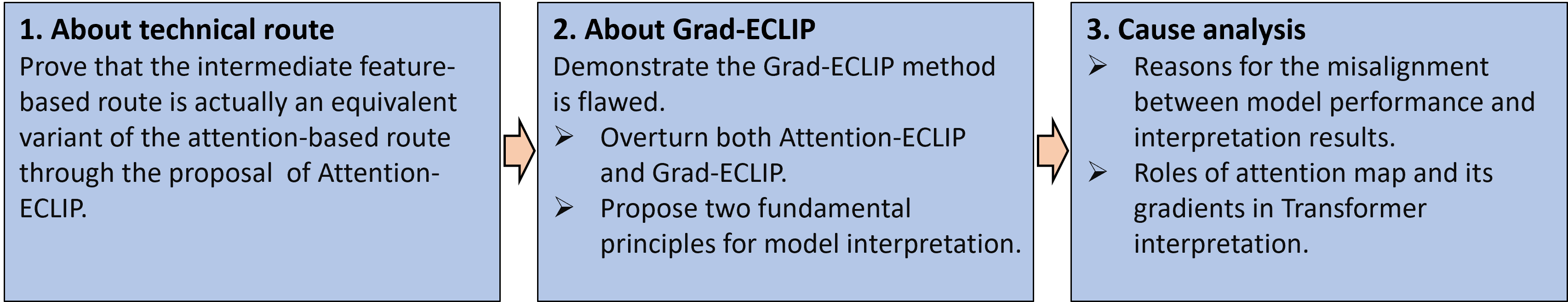}
    \caption{The overall idea of this work.}
    \label{silu}
    \vspace{-0.7cm}

    \end{figure*}

Transformer was first proposed and applied in the field of natural language processing \cite{1}, and later successfully applied to computer vision \cite{6,7} as well as audio field \cite{12,13}, multimodal field \cite{21,22,23,CLIP} and other crossing fields like pharmaceuticals \cite{28,29,30}. And it has spawned a series of large models \cite{38,41,42}, promoting the further development of artificial intelligence. 

Among numerous multimodal models, the Contrastive Language-Image Pre-training (CLIP) \cite{CLIP} stands out for its exceptional performance. Through the process of learning representations that align caption text with its corresponding image, CLIP model has presented a straightforward yet highly effective dual-encoder pre-training framework, facilitating interaction between natural language processing and computer vision. Leveraging zero-shot learning and fine-tuning techniques, CLIP has markedly enhanced performance across a range of downstream tasks, including classification \cite{add1_1}, retrieval \cite{add2_1}, and segmentation \cite{add3_1,add3_2}. Drawing inspiration from CLIP, researchers have further advanced multi-modal pre-training by exploring diverse avenues, such as integrating vision-language understanding and generation \cite{add4_1}, designing prompts \cite{add5_1}, and implementing region-aware enhancements \cite{add6_1}.

With the significant development of artificial intelligence driven by Transformer, people's concerns have arisen. These concerns mainly stem from the unexplainable characteristics of AI, which may bring unknown problems, such as algorithmic safety issues in AI healthcare and autonomous driving, and algorithmic fairness issues in treating different groups. Therefore, relevant regulations were born. The AI Act established by the European Commission outlines rules and norms to govern the responsible use and deployment of AI, and proposed the three elements of trustworthy AI: legality, ethics, robustness. EU's General Data Protection Regulation ensures users' right to total transparency regarding decisions made by automated systems. The AI strategic plan established by America Select Committee on Artificial Intelligence of National Science and Technology Council also emphasize the need to address the potential adverse effects of AI through model interpretation. Model interpretation has become imperative to ensure AI's trustworthiness, as well as its widespread promotion and application.

Attention-based methods, such as Generic Attention-model Explainability (GAE) \cite{48}, treat model attention as the most crucial feature and leverage the propagation of attention across different attention layers for model interpretation. Among intermediate feature-based methods, the Gradient-based Visual Explanation Method for CLIP (Grad-ECLIP) \cite{gradeclip}, published at the International Conference on Machine Learning (ICML) 2024, is the most representative. Intermediate feature-based methods primarily utilize intermediate features to measure the contribution of the location where the feature resides to the model's output.

Grad-ECLIP served as the catalyst for this research, as it introduced a novel approach to Transformer interpretation and garnered widespread attention following its publication at ICML 2024. When submitting our work related to attention-based model interpretation, we were repeatedly asked to compare it with Grad-ECLIP. However, we observed that reviewers did not fully comprehend the distinctions and connections between Grad-ECLIP, an intermediate feature-based method, and attention-based methods. Grad-ECLIP is a flawed algorithm that garnered significant attention after its publication at ICML 2024, which subsequently impacted the reception of our related work. Therefore, we deemed it highly necessary to conduct a detailed study on this matter.

\textbf{The overall idea of this work is shown in Figure \ref{silu}.}

\textbf{Our contributions mainly consist of the following three aspects:}
\begin{itemize}
    \item We propose an attention-based method (Attention-ECLIP) equivalent and more concise to Grad-ECLIP, achieving a unification of the two distinct technical routes.
    \item We overturn both Grad-ECLIP and Attention-ECLIP, and propose two fundamental principles for model interpretation.
    \item We explain the phenomenon of misalignment between model performance and interpretation results, and elucidate the roles of attention map and its gradients in Transformer interpretation.
\end{itemize}

\section{Related Work}

\subsection{Contrastive Language-Image Pre-training (CLIP).}

CLIP \cite{CLIP} is a model that integrates Transformer architecture with contrastive learning. It projects text and image features extracted by Transformer blocks into the same space and then calculates the similarity between the information of these two modalities within that space.
Among numerous multimodal models, the Contrastive Language-Image Pre-training (CLIP) \cite{CLIP} stands out for its exceptional performance. Through the process of learning representations that align caption text with its corresponding image, CLIP model has presented a straightforward yet highly effective dual-encoder pre-training framework, facilitating interaction between natural language processing and computer vision. Leveraging zero-shot learning and fine-tuning techniques, CLIP has markedly enhanced performance across a range of downstream tasks, including classification \cite{add1_1}, retrieval \cite{add2_1}, and segmentation \cite{add3_1,add3_2}. Drawing inspiration from CLIP, researchers have further advanced multi-modal pre-training by exploring diverse avenues, such as integrating vision-language understanding and generation \cite{add4_1}, designing prompts \cite{add5_1}, and implementing region-aware enhancements \cite{add6_1}.

This paper takes the CLIP model as the object model to interpret, primarily because Grad-ECLIP, which is based on intermediate features, is specifically designed for CLIP.

\subsection{Transformer interpretation methods.}

The interpretation methods of the Transformer model mainly utilize its unique attention structure \cite{1,6}, or combine with the interpretation methods such as GradCAM \cite{51}, LayerCAM \cite{52}, LRP \cite{53}, etc. Vaswani et al. \cite{1} applied the attention of partial layers and partial heads to interpret the intrinsic mechanism of Transformer when they first proposed it, and found that different heads perform different tasks. Multi-head mechanism has become a very important issue in model interpretation. Michel et al. \cite{57} obtained the same conclusion that different heads perform different tasks and contributed differently, and therefore proposed that pruning the unimportant heads has little impact to the model. Considering information originating from different tokens gets increasingly mixed, making attention weights unreliable as interpretation probes, Abnar et al. \cite{58} proposed attention rollout and attention flow to quantify the flow of information through self-attention. And Dosovitskiy et al. \cite{6} applied attention rollout to compute maps of the attention from the output token to the input space when they first proposed Vision Transformer (ViT). Chefer et al. \cite{49} introduced Transformer Attribution (T-Attr) integrating scores throughout the attention graph, by incorporating both LRP-based relevancy and gradient information, in a way that iteratively removes the negative contributions. Chefer et al. \cite{48} introduced Generic Attention-model Explainability (GAE), which combines gradient with multi-head attention maps, and then performing attention rollout. Yuan et al. \cite{yuan2021explaining} interpret information flow inside Vision Transformers using Markov Chain (TAM). Barkan et al. \cite{47} propose Deep Integrated Explanations (DIX), generates explanation maps by GAE and Integrated Gradient. Xie et al. \cite{DBLP:conf/ijcai/Xie0CZ23} propose ViT-CX based on token embeddings, rather than attentions paid to them, and their causal impacts on the model output. Englebert et al. \cite{DBLP:conf/iccvw/EnglebertSNMSCV23} propose Transformer Input Sampling (TIS) a perturbation-based explainability method for Vision Transformers, which computes a saliency map based on perturbations induced by a sampling of the input tokens. Chen et al. \cite{46} propose Beyond Intuition Method (BI) to approximate token contributions inside Transformers, in order to solve  the ambiguity of the expression formulation which can lead to an accumulation of error. Zhao et al. \cite{gradeclip} propose Grad-ECLIP to interpret Contrastive Language-Image Pre-training (CLIP).

Among the aforementioned Transformer interpretation methods, we primarily categorize them into four types: perturbation-based, relevance-based, attention-based, and intermediate feature-based. Perturbation-based methods include ViT-CX and TIS. The relevance-based approach is represented by T-Attr, which allocates relevance scores from the output back to the input iteratively. Attention-based methods encompass Attention Rollout, GAE, TAM, DIX, and BI; these methods generally rely on the propagation of attention across layers to obtain importance scores. The intermediate feature-based method is Grad-ECLIP.

\section{Method}

The impetus for this work stems from the fact that Grad-ECLIP, published at ICML 2024, introduced a novel approach to Transformer interpretation, which has garnered widespread attention. Our various attention-based works have all been requested to be compared with Grad-ECLIP. However, we found that reviewers did not have a profound understanding of the two distinct technical routes—attention-based and intermediate feature-based routes. We believe it is highly necessary to conduct an in-depth study on this matter.

The attention-based method, exemplified by GAE, achieves Transformer interpretation by calculating the iterative propagation process of attention. The intermediate feature-based method is Grad-ECLIP, which is primarily applied to the CLIP model. Therefore, we also mainly conducted our experiments based on the CLIP model.

\begin{figure}[]
    \centering
    \includegraphics[width=0.9\columnwidth]{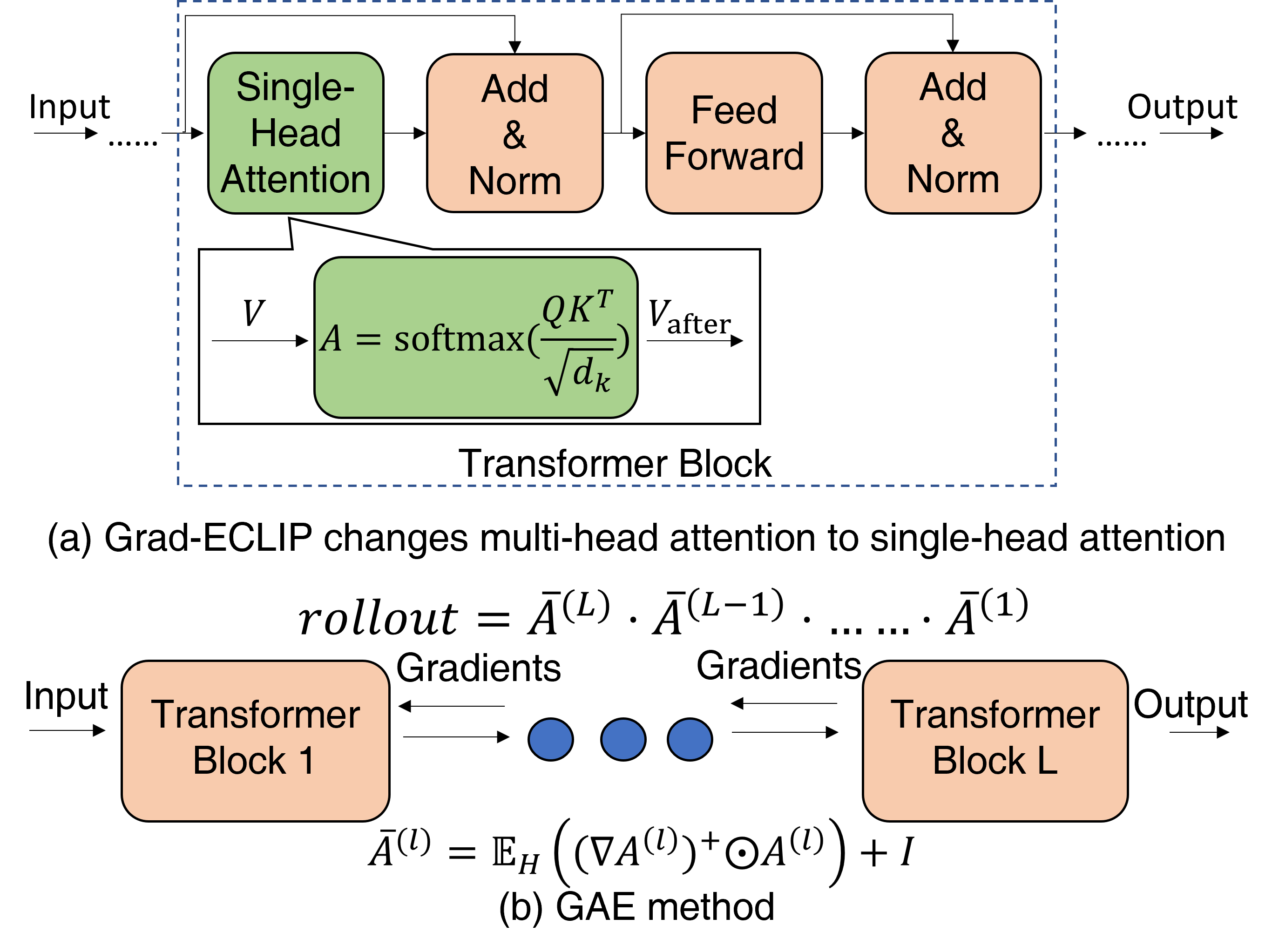} 
    \caption{Grad-ECLIP and GAE. ($Q$, $K$, $V$ are Query, Key, Value matrices; $A$ is the attention map; $\bar{A}$ is the mean attention map across multi-head attention maps; $\frac{1}{\sqrt{d_k}}$ is the scaling factor.)}
    \label{myfig1}
    \vspace{-0.7cm}

    \end{figure}

\subsection{Grad-ECLIP}
The computational process of Grad-ECLIP is illustrated in Figure \ref{myfig1} (a), which depicts a specific attention layer within the model. Initially, the model is adjusted to convert the relevant attention layer (for CLIP, the last attention layer of the image encoder and the last eight attention layers of the text encoder are utilized for explanation) from a multi-head configuration to a single-head one. Subsequently, the interpretation results obtained from the adjusted single-head model are treated as the interpretation results of the original model. 

The process is divided into two stages. \textbf{In the first stage}, the $Output$ in Figure \ref{myfig1} (a) represents the output of the entire model—for the CLIP model, it is the final cosine similarity between the text and image. Next, the $loss$ is set to this $output$, and the gradient with respect to $V_{after_{cls}}$ is computed. This gradient is then used as the weight for each token feature in $V$ to calculate the weighted value of each token feature by summing up, which serves as the first component of the interpretation result.

\begin{equation}\label{eqadd1}
    \mathrm{Gstage}1_{i}=\sum_{\mathrm{dim}}(\frac{\partial Output}{\partial V_{after_{cls}}}\odot V_{i})
\end{equation}
where $\odot$ is the Hadamard product, $dim$ represents the dimension of intermediate token features, $i$ represents the $i_{th}$ token.

\textbf{In the second stage}, because the author believes that the attention map processed by softmax is too sparse, the attention map is recalculated as follows.
\begin{equation}\label{eq1}
    {Gstage}2_{i}=w_i=\Phi(\Phi_1(q_{cls})(\Phi_1(k_i))^\mathsf{T})
\end{equation}
where $\Phi_1$ represents ${torch.nn.functional.normalize}$, $\Phi$ represents the Min-Max Normalization. The algorithm process can be referred to in Algorithm \ref{alg1}.

\begin{algorithm}[]
    \caption{Attention map adjustment by Grad-ECLIP}
    \label{alg1}
    \textbf{Input}: $q_{\text{out}}, k_{\text{out}}$
    \textbf{Output}: $w_{qk}$
    \begin{algorithmic}[1]
    \STATE $q_{\text{cls}} \gets q_{\text{out}}[:1,0,:]$;
    $k_{\text{token}} \gets k_{\text{out}}[1:, 0, :]$
    \STATE $q_{\text{cls}} \gets \text{normalize}(q_{\text{cls}})$;
    $k_{\text{token}} \gets \text{normalize}(k_{\text{token}})$
    \STATE $w_{qk} \gets \sum(q_{\text{cls}} \odot k_{\text{token}})$\;
    \STATE $w_{qk} \gets (w_{qk} - \min(w_{qk})) / (\max(w_{qk}) - \min(w_{qk}))$\;
    \STATE \textbf{return } {$w_{qk}$}
    \end{algorithmic}
    \end{algorithm}

And then, the $Gstage2_i$ is used as the weight for the score of the $i_{th}$ token obtained in the first stage to get the final score.

\begin{equation}\label{eqadd2}
    Gscore_i=Gstage1_i \odot Gstage2_i
\end{equation}

\subsection{GAE}
The computational process of GAE is illustrated in Figure \ref{myfig1} (b). GAE uses the positive gradient of the attention map as the weight to merge multi-head attention into single-head attention, and then follows the calculation procedure of attention rollout to derive the final interpretation score. The detailed computational steps are described below.
\begin{equation}\label{eq2}
    \bar{A}^{(l)}=I+\mathbb{E}_H((\nabla A^{(l)})^{+}\odot A^{(l)})
\end{equation}
\begin{equation}\label{eq3}
    rollout=\bar{A}^{(L)}\cdot\bar{A}^{(L-1)}\cdot...\cdot\bar{A}^{(2)}\cdot\bar{A}^{(1)}
\end{equation}
\begin{equation}\label{eq4}
    C=rollout[0,1:]
\end{equation}
Where the identity matrix $I$ is added to represent the contribution of the residual connection part, $H$ represents number of heads, $L$ represents number of attention layers, $l$ represents the $l_{th}$ layer, $L_{th}$ layer represents the last layer. Because the final input to the MLP head is the $cls$ token, attention rollout takes the $rollout$ value of $cls$ token as $C$, the overall contribution value from the perspective of input or overall attention value from the perspective of model, $\mathbb{E}$ represents the operation of taking the mean, $\cdot$ represents matrix multiplication, $\odot$ is the Hadamard product, $\nabla$ represents gradients.

\subsection{Attention-ECLIP}
Next, we implement Gradient-based visual Explanation method for CLIP (Grad-ECLIP) based on attention, and thus name our method Attention-based Visual Explanation for CLIP (Attention-ECLIP).

Attention-ECLIP handles the model in the same way as Grad-ECLIP does—it sets the target attention layer to a single-head attention and does not take residual connections into account. 

A key approach in GAE for interpreting Transformers is to utilize gradients as weights for the attention map, denoted as $\mathbb{E}_H((\nabla A)^{+}\odot A)$. Since we simplify the multi-head attention in the target attention layer to single-head attention as Grad-ECLIP does, this calculation process becomes equivalent to $(\nabla A)^{+}\odot A$. In alignment with Grad-ECLIP, we adopt the complete gradient, transforming the calculation process into $\nabla A\odot A$. 

We believe that, within the same information pathway, identical model results can be achieved whether viewed from the perspective of attention or intermediate features. In other words, Attention-ECLIP based on attention and Grad-ECLIP based on intermediate features can achieve the same effect.

Given that the second stage of Grad-ECLIP involves the use of processed attention maps, denoted as $Gstage2$ (Equation \ref{eq1}), and this part does not pertain to the choice of technical route, we temporarily employ the identical second stage.

\begin{equation}\label{eq1_1}
    {Astage}2=Gstage2=w
\end{equation}

Then,
\begin{equation}\label{eq1_2}
    {Astage}1={\nabla A}_{cls}
\end{equation}
Then, the score of $i_{th}$ token is as follow,
\begin{equation}\label{eqadd1_3}
    Ascore_i={{\nabla A}_{cls}}_i \odot w_i
\end{equation}

In subsequent experiments, we will verify the equivalence between Attention-ECLIP without $w_i$ and Grad-ECLIP without $w_i$, as well as between Attention-ECLIP and Grad-ECLIP.

\textbf{Please note:}
\begin{itemize}
    \item We do not endorse replacing the attention map A with $w$. We only adopt this approach temporarily because $w$ does not affect the choice of technical route. We provide detailed discussions on this matter in Subsection \ref{wi}.
    \item We will overturn both Grad-ECLIP and Attention-ECLIP in Subsection \ref{tuifan}, and propose two fundamental principles that model interpretation should adhere to in Subsection \ref{liangyuanze}.
    \item We will explain the phenomenon of misalignment between model performance and interpretation results, and elucidate the roles of attention map and its gradients in Transformer interpretation in Subsection \ref{liangyaosu}.
\end{itemize}

\section{Experiments}

\subsection{Formal Derivation of Equivalence.}

Both Grad-ECLIP and Attention-ECLIP set the multi-head attention of a certain layer to single-head attention as shown in Figure \ref{myfig1} (a). 

In the following formula derivation, $\cdot$ represents matrix multiplication, $\odot$ represents the Hadamard product, $L$ represents $loss = output$, $tr$ represents the trace, and $cls$ represents the row where the $cls$ token is located in the attention map or the intermediate feature of the $cls$ token.

\begin{equation}\label{eqadd3}
    V_{after}=A \cdot V
\end{equation}
The derivative of L with respect to $V_{after}$ is
\begin{equation}\label{eqadd4}
    dL=tr\left(\left(\frac{\partial L}{\partial V_{after}}\right)^T\cdot dV_{after}\right)
\end{equation}
Because $dV_{after}=(dA) \cdot V$, so
\begin{equation}\label{eqadd5}
    dL=tr\left(\left(\frac{\partial L}{\partial V_{after}}\right)^T\cdot dA \cdot V\right)
\end{equation}
Utilizing the cyclic property of the trace,
\begin{equation}\label{eqadd6}
    dL=tr\left(V \cdot \left(\frac{\partial L}{\partial V_{after}}\right)^T\cdot dA\right)
\end{equation}
And the derivative of L with respect to A is,
\begin{equation}\label{eqadd7}
    dL=tr\left(\left(\frac{\partial L}{\partial A}\right)^T\cdot dA\right)
\end{equation}
By comparing Equation \ref{eqadd6} and Equation \ref{eqadd7}, we can obtain
\begin{equation}\label{eqadd8}
    \left(\frac{\partial L}{\partial A}\right)^T=V\cdot \left(\frac{\partial L}{\partial V_{after}}\right)^T
\end{equation}
Furthermore,
\begin{equation}\label{eqadd9}
    \frac{\partial L}{\partial A}=\frac{\partial L}{\partial V_{after}}\cdot V^T
\end{equation}
At this point, we only consider the interpretation corresponding to the $cls$ token.
\begin{equation}\label{eqadd10}
    \frac{\partial L}{\partial A_{cls}}=\frac{\partial L}{\partial (V_{after})_{cls}}\cdot V^T
\end{equation}
We will now proceed to provide an explanation for the $i_{th}$ token.
\begin{equation}\label{eqadd11}
    \frac{\partial L}{\partial (A_{cls})_i}=\frac{\partial L}{\partial (V_{after})_{cls}}\cdot (V_i)^T=\sum_{\mathrm{dim}}(\frac{\partial Output}{\partial (V_{after})_{cls}}\odot V_{i})
\end{equation}
That is to say,
\begin{equation}\label{eqadd12}
    \nabla {A_{cls}}_i=\sum_{\mathrm{dim}}(\frac{\partial Output}{\partial (V_{after})_{cls}}\odot V_{i})
\end{equation}

Up to this point, we have formally proven through derivation that the first stage of Grad-ECLIP and Attention-ECLIP are completely equivalent.

\subsection{Experimental Verification of Equivalence.} {\label{sec:ex veri}}

We have already completed the formal derivation verification. To make it more convincing, we will now conduct experimental validation. The purpose of these experiments is to verify the equivalence. Since the setting of the target attention layer does not affect the  verification of equivalence, we thus set the target layer as the last attention layer for both text and image interpretation.
We adopt two approaches for analysis: one is case demonstration, which is used to visually present the results (Figure \ref{myfig2}), and the other is batch quantitative analysis (Table \ref{table2} and \ref{table3} in Appendix \ref{buchongdingliang}), which is employed to eliminate the one-sidedness of case demonstration.

In batch quantitative experiments, we need to utilize certain metrics to describe the experimental results. For image interpretation, we adopt the Grad-ECLIP experimental approach. Specifically, we conduct insertion (insert image pixels into a blank image in descending order of importance, with a step size of 10\%) and deletion (remove and replace important pixels with random values in descending order of importance, with a step size of 10\%) experiments \cite{gradeclip} on images and then calculate the model recognition rates of labels (Top1). The experimental results are presented in Table \ref{table2} in Appendix \ref{buchongdingliang}. For text interpretation, we perform text deletion experiments, removing one token at a time for a total of five deletions in descending order of importance, and then calculate the cosine similarity. The experimental results are shown in the Table \ref{table3} in Appendix \ref{buchongdingliang}.

The dataset for the image quantification experiments is ImageNet \cite{62}, and the dataset for the text quantification experiments is MS COCO \cite{coco}. Both datasets consist of 1,000 samples, which are uniformly selected from their respective original datasets.

\begin{figure*}[]
    \centering
    \includegraphics[width=1.5\columnwidth]{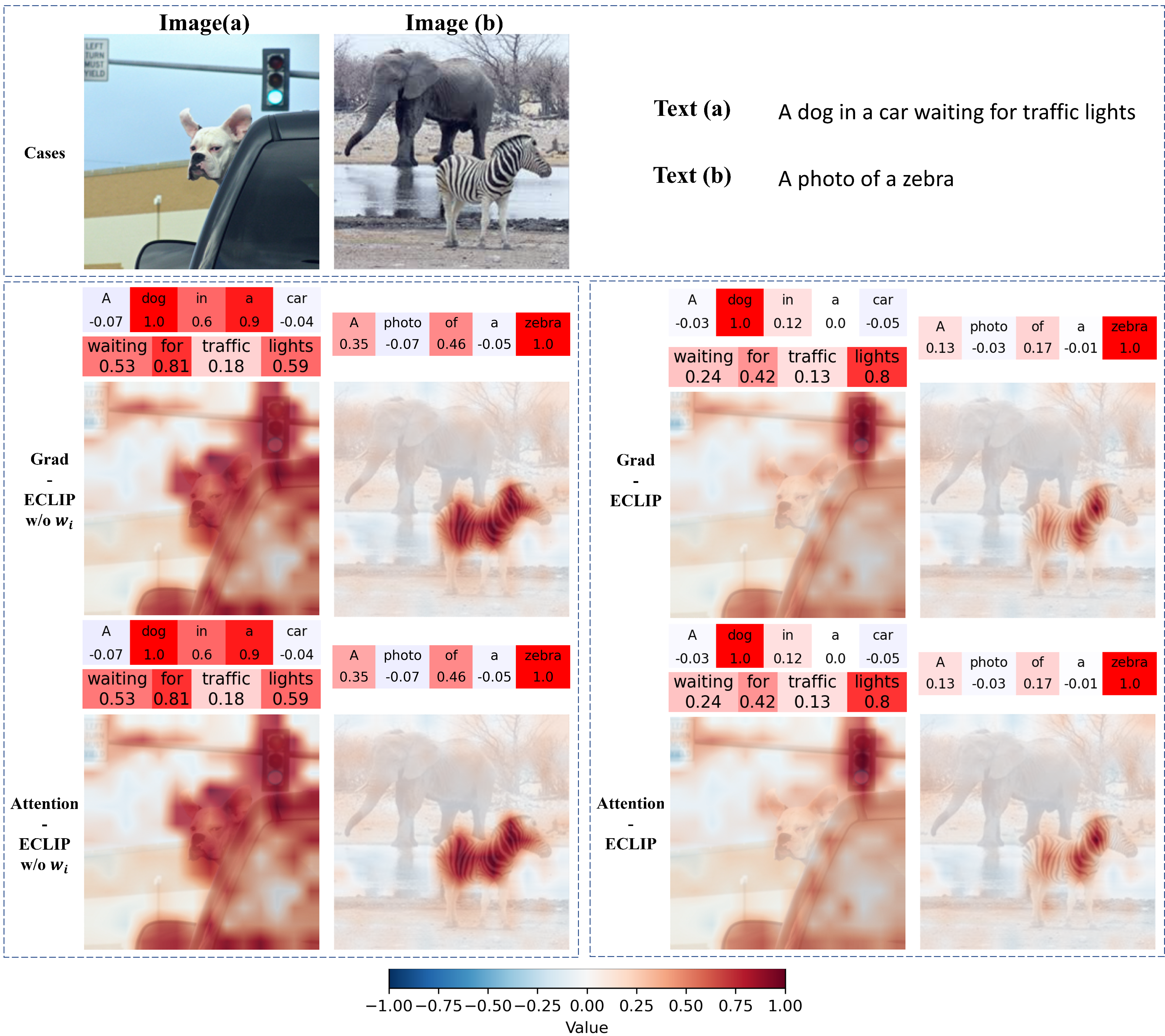} 
    \caption{Interpretation results of Grad-ECLIP w/o $w_i$, Attention-ECLIP w/o $w_i$, Grad-ECLIP and Attention-ECLIP.}
    \label{myfig2}
    \vspace{-0.4cm}

    \end{figure*}

From the aforementioned experimental results, it can be observed that Grad-ECLIP and Attention-ECLIP are equivalent in both qualitative and quantitative experiments. The minor discrepancies in the quantitative experiments stem from slight differences in algorithm implementation or numerical calculation, floating point error, and random processes. Readers can verify more experimental results through the code that we will subsequently make publicly available.

Compared to Grad-ECLIP, Attention-ECLIP leverages attention maps instead of intermediate features, reducing gradient computation requirements and eliminating a summation operation. This design renders it both more concise and computationally efficient.

\textbf{Up to now, we have confirmed that the two methods are equivalent. If you agree with the concept behind Grad-ECLIP, our current contribution lies in achieving the same effects as Grad-ECLIP in a more concise and efficient manner based on attention, and proving that the intermediate feature-based route is actually an equivalent variant of the attention-based route . In other words, we have achieved a unification of the two technical routes: the intermediate feature-based approach and the attention-based approach. However, this work has prompted us to engage in deeper reflections as follows.}

\subsection{Is Grad-ECLIP reasonable?}\label{tuifan}

We have demonstrated that Grad-ECLIP and Attention-ECLIP are equivalent. Here, we aim to discuss whether the technical approach represented by these two methods is reasonable. This technical approach has two key characteristics: First, it forcibly converts the target multi-head attention layer into a single-head attention layer, altering the model architecture and using the interpretation results from the modified model as interpretation for the original model. Second, it involves recalculating the attention map component, which corresponds to the second stage of the algorithm. In short, we hold an opposing view towards Grad-ECLIP.

\subsubsection{Two fundamental principles for model interpretation}\label{liangyuanze}

We believe that the two principles proposed in this section represent the most fundamental requirements for model interpretation, even serving as implicit defaults within the field. Our presentation here is more accurately described as explicitly highlighting these principles.

Both Grad-ECLIP and Attention-ECLIP forcibly alter the model structure and uses the interpretation results of the modified model as those of the original model, which we consider unreasonable. As shown in Table \ref{table4}, taking the two cases in Figure \ref{myfig2} as examples, we combine different texts and images for experiments to compare the outputs of the original model and the modified model respectively. We find that the performance of the modified model significantly declines, and it cannot accurately determine the associations between text and images. \textbf{More experimental evidence is provided in the Appendix \ref{comp}.} Using the interpretation results of an inferior (modified) model to represent those of the original model is clearly unreasonable. 

Here, we propose the first principle that model interpretation should adhere to: \textbf{model interpretation should be faithful to the original model}.

\textbf{Please note: all Grad-ECLIP interpretations of the CLIP model are actually those of the modified CLIP model.}

This is actually a logical fallacy: using the interpretation results of the modified model as those of the original model. Then, how are we supposed to interpret the modified model itself? The modified model can be directly interpreted, which means the original model and the modified model share one and the same exact interpretation result. For two models with different structures and performance to share an identical interpretation result is inherently absurd.

\begin{table}[]
    \centering
    \caption{Performance of the original CLIP and modified CLIP (with the last attention layer of the image encoder is modified to single-head attention).((a) and (b) represent cases in Figure \ref{myfig2})}
    \label{table4}
    \begin{tabular}{@{}lllll@{}}
    \toprule
    Image/Text    & (a)/(a) & (b)/(b) & (b)/(a) & (a)/(b) \\ \midrule
    Original CLIP & 29.10    & 35.06   & 16.26   & 14.11   \\
    Modified CLIP & 17.55   & 17.50    & 22.53   & 15.71   \\ \bottomrule
    \end{tabular}
    \vspace{-0.2cm}

    \end{table}

Now we interpret the modified model. Another issue can also be observed from Table \ref{table4}: the output of the modified model indicates that the model perceives the association between the image and text as very weak, yet the interpretation results are surprisingly good (Figure \ref{myfig2}). This is clearly unreasonable, because an important function of model interpretation is to reflect model performance. Here, we propose the second principle that model interpretation should follow: \textbf{the model interpretation results should align with the model's performance}.

\subsubsection{Additional Explanation on the Principle of Fidelity}

We have noticed that some methods also achieve model interpretation by constructing a surrogate model, such as LIME\cite{DBLP:conf/kdd/Ribeiro0G16}, TreeView\cite{DBLP:journals/corr/ThiagarajanKSR16}, CAM\cite{50},model extraction\cite{DBLP:journals/corr/BastaniKB17a}. However, these surrogate models are all retrained to ensure that their performance is sufficiently similar to that of the original model, thereby guaranteeing fidelity. Grad-ECLIP is starkly different from the aforementioned methods. It directly applies the original model's parameters to the surrogate model without retraining, resulting in a significant performance disparity between the two models.

Using the interpretation of a surrogate model as that of the original model has certain drawbacks. For instance, even if the performance of a surrogate model is close to that of the original model, if their reasoning processes differ, the interpretation results obtained may carry a risk of misleading\cite{DBLP:conf/xai/MariottiSA23}. Even disregarding these risks, differences between the surrogate and original models will always persist. From a developmental perspective, we still prefer that model interpretation should remain faithful to the original model.

Therefore, our interpretation of fidelity is that it is best not to alter the model structure. If the model structure must be changed to create a surrogate model, it is essential to ensure that the original and surrogate models are sufficiently similar.

\subsubsection{Then why can the inferior modified CLIP model yield good interpretation results?}\label{liangyaosu}

\begin{figure}[]
    \centering
    \includegraphics[width=1\columnwidth]{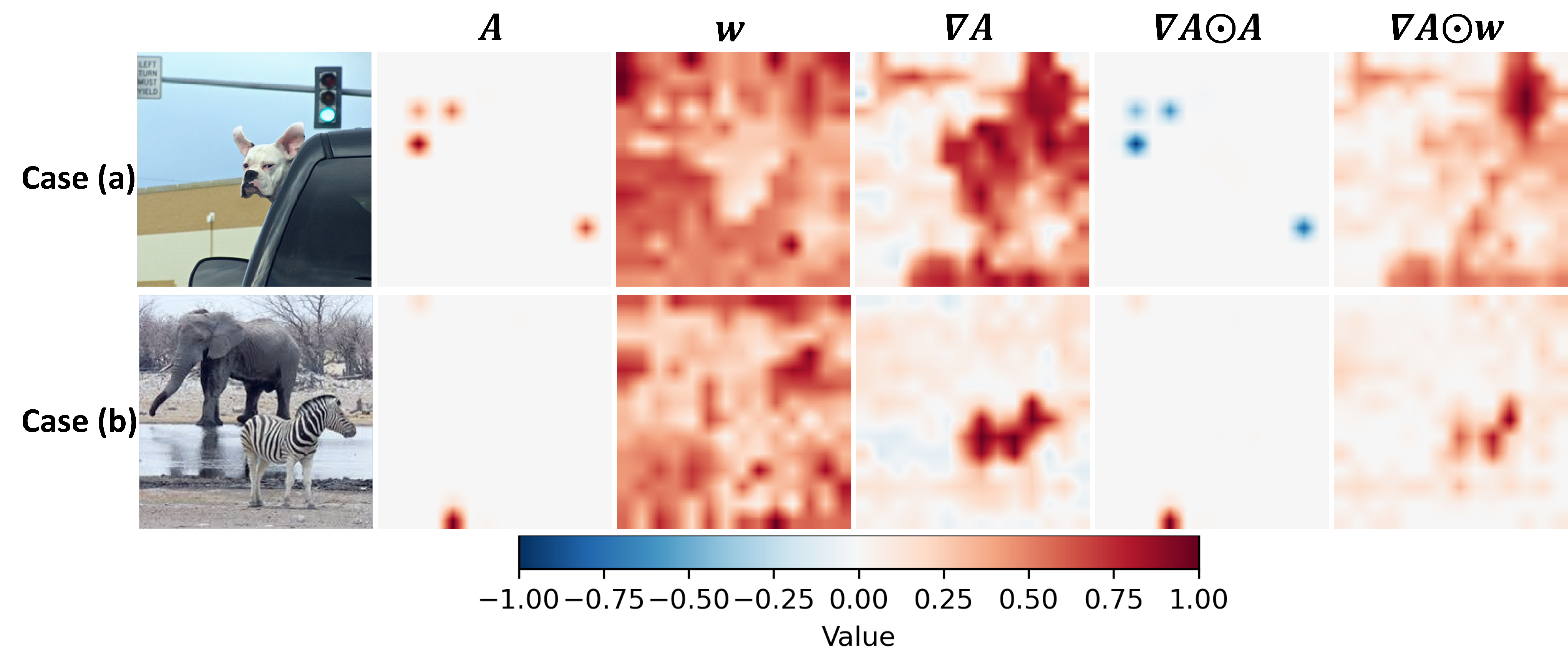}

    \caption{Comparison of Interpretation Effects Among Different Components.($\nabla A\odot w$ is equivalent to Grad-ECLIP.)}
    \label{10051}
    \vspace{-0.4cm}

    \end{figure}

\begin{figure*}[]
    \centering
    \includegraphics[width=1.5\columnwidth]{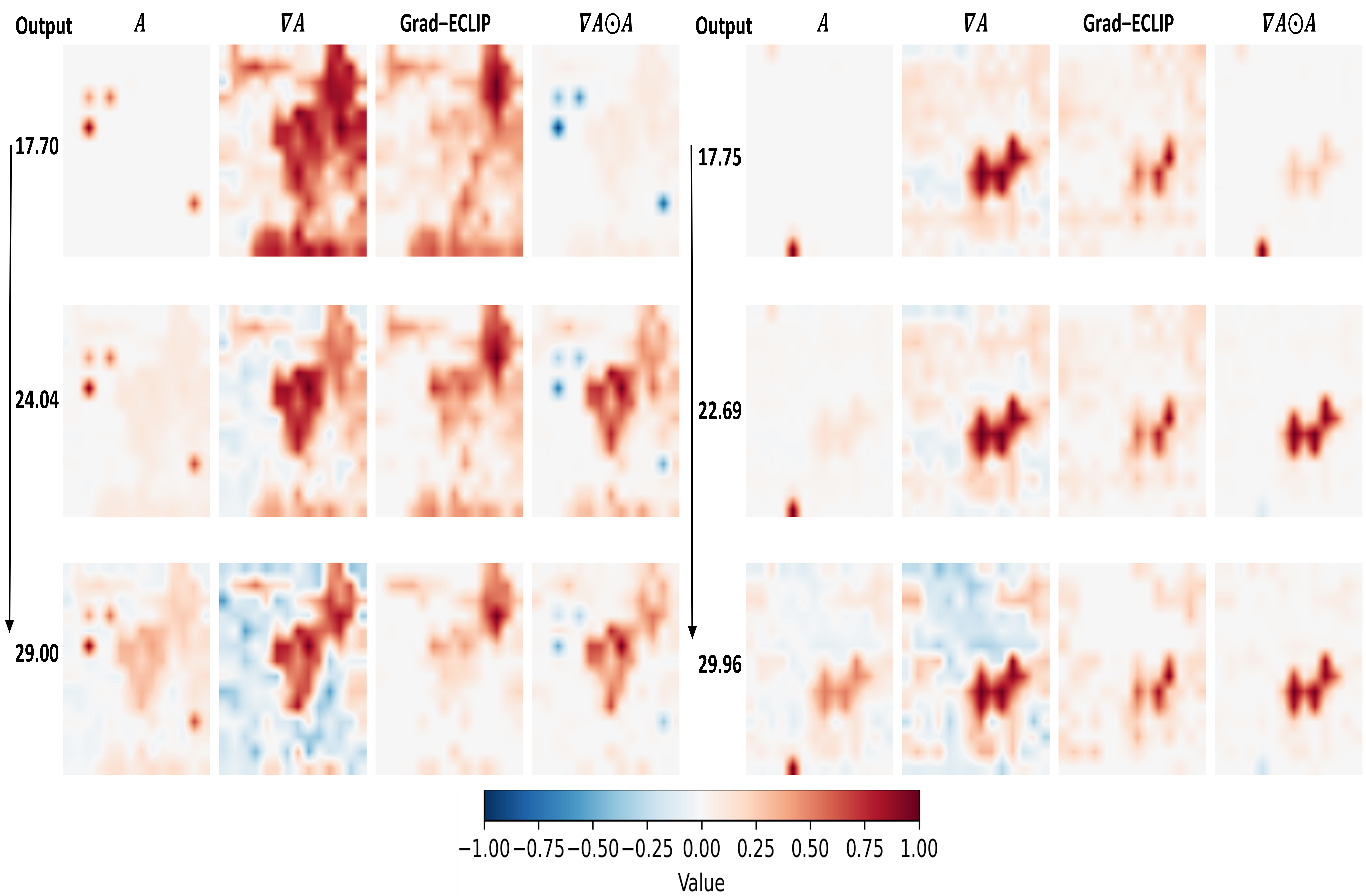}
    \caption{Experiment on updating attention map.(As $A$ undergoes updates, the model's performance continues to improve; the interpretation results from $ \nabla A\odot A$ align with the model's performance well. Because the computation of $w$ is not directly associated with $A$, therefore, as $A$ updates, $w$ remains unchanged, and $w$ is shown in Figure \ref{10051}.)}

    \label{myfig3}
    \vspace{-0.5cm}

    \end{figure*}

Since Attention-ECLIP has been proven equivalent to Grad-ECLIP and is relatively more concise, we take Attention-ECLIP as an example for analysis here. Model interpretation involves two main components: one is the attention map (Grad-ECLIP has readjusted its computational process as shown in Equation \ref{eq1} and Algorithm \ref{alg1}), and the other is the gradient of the attention map (equivalent to Equation \ref{eqadd1} of Grad-ECLIP).

The approach of using gradients as attention weights was first proposed in the GAE algorithm, and subsequent algorithms such as TAM, BI, and DIX have all adopted this method. However, none of these algorithms provided an explanation for it. Here, we discuss the role of attention map and its gradients from two perspectives: linear approximation and model optimization.

Attention map reflects the degree to which the current attention layer focuses on different tokens. However, a model's inference does not rely solely on which tokens the attention mechanism prioritizes; it also depends on different decision-making mechanisms, such as the classifier in ViT or the relevance calculation component in CLIP. Therefore, relying solely on a model's attention cannot fully explain the basis for its inference.

\textbf{From the perspective of linear approximation}: Gradients reflect the sensitivity of the model's output to changes in variables, indicating the importance of the current variable to the output. Thus, using gradients as weights for attention map measures the contribution of current attention to the model's output. Formally, this is a linear approximation under the assumption of local linearity, where the contribution is approximated as $A \odot \frac{\partial Output}{\partial A}$.

\textbf{From the perspective of model optimization}: Gradients represent the magnitude and direction of variable optimization. For example, in Figure \ref{10051} and \ref{myfig3} (Figure \ref{10051} is the result of modified model, Figure \ref{myfig3} is the result of modified model after updating the attention map using gradients), gradients indicate the regions where attention should be more focused. Attention map, on the other hand, reflects the regions currently being attended to. Grad-ECLIP applies smoothing to the attention map (Equation \ref{eq1}, Subsubsection \ref{wi}), causing the interpretation results to be overly reliant on gradient information, that is, highlighting the regions the model should ideally attend to. Although the obtained results appear favorable, but fail to align with the model's performance.

We conducted the experiment shown in Figure \ref{myfig3}. Taking image interpretation as an example, we considered the last layer of attention in the image encoder as the target attention layer, set it as single-head attention, and made the attention map $A$ the only updatable parameter of the model. Subsequently, set $loss=-output$, where $output$ is the value of the model output. Then we continuously update the attention map and observe the change of model output, model attention and other components. The interpretation results from $A\odot \nabla A$ aligned well with the model's performance.

In summary, by using the adjusted underperforming model instead of the original model for interpretation, and downplaying the actual attention distribution of the underperforming model to highlight the ideal attention distribution as the interpretation result attributed to the original model, the Grad-CLIP method ultimately appears to provide a seemingly flawless interpretation for the original model. However, it is incorrect.

This experiment also demonstrates that the attention map $A$ holds significant interpretative value, and we do not endorse the adjustments made to the attention map by the Grad-ECLIP. We will discuss this in the following Subsection \ref{wi}.

\subsubsection{Attention map adjustment by Grad-ECLIP}\label{wi}

The second stage of Grad-ECLIP recalculates the attention map because the softmax self-attention function is extremely sparse. The computational process is illustrated in Algorithm \ref{alg1} and Equation \ref{eq1}. However, this process may not only alter the sparsity but also change the relative importance relationships among different tokens. We demonstrate this with a simple example below.

Let:
 $q_{cls} = [1, 0]$;
 $k_1 = [0.8, 0.6]$;
 $k_2 = [1.414, 1.414]$

According to the attention map calculation rules, before applying softmax, $q_{cls}k_{1}^T = 0.8$ and $q_{cls}k_{2}^T = 1.414$. Since softmax preserves relative magnitudes, the final result will still be $A_{q_{cls}k_{1}} < A_{q_{cls}k_{2}}$.

However, under the calculation method in the second stage of Grad-ECLIP (Algorithm \ref{alg1}), the final results are $w_{q_{cls}k_{1}} \approx 8.60$ and $w_{q_{cls}k_{2}} = 0$, that is $w_{q_{cls}k_{1}} > w_{q_{cls}k_{2}}$. This operation may change the relative magnitudes between different tokens, which can also be verified in Figure \ref{10051}.

Most importantly, the second stage of Grad-ECLIP fails to ensure that the final results adhere to the consistency principle, as the original attention map does, refer to Subsection \ref{liangyaosu}.

\section{Application of the consistency principle}

Part of another study of ours involves the application of the consistency principle we proposed (the model interpretation results should align with the model performance). We utilized this principle to identify two erroneous algorithms, TAM\cite{yuan2021explaining} and BI\cite{46}. This study is about to be made public, with the paper titled "\textbf{Integrated Gradients in Transformer Interpretation}".

\section{Reproduction}
The code will be made public later.

\section{Conclusion}
This work might be suspected of being offensive or targeted, so it is necessary for us to clarify our stance. This research originated from the fact that reviewers for several of our previous works requested comparisons with Grad-ECLIP. However, we found that even the reviewers struggled to fully grasp the distinctions between these two technical approaches: intermediate feature-based and attention-based. Grad-ECLIP, as a notable achievement published at a prestigious conference like ICML 2024, has garnered significant attention, and its technical approach intrigued us. The flaws of Grad-ECLIP, after gaining recognition at ICML, subsequently impacted the reception of our related work. We deemed it essential to conduct a comprehensive analysis, which led to the creation of this paper. We do not deny the contributions of Grad-ECLIP. In fact, its emergence has prompted us to clarify the principles that model interpretation should adhere to, spurred us to explore the roles of attention map and its gradients in Transformer interpretation, and encouraged us to investigate the similarities and differences between the two technical routes. We hope to avoid the emergence of similar issues by applying the two principles proposed in this paper. We also hope that our unification of the two technical can provide assistance for subsequent research. Furthermore, we hope that our interpretations of the gradient and attention map can inspire new explanatory methods.

\section*{Impact Statement}

There are many potential societal consequences of our work, none
which we feel must be specifically highlighted here.


\bibliography{danmotai}

\begin{thebibliography}{43}
\providecommand{\natexlab}[1]{#1}
\providecommand{\url}[1]{\texttt{#1}}
\expandafter\ifx\csname urlstyle\endcsname\relax
  \providecommand{\doi}[1]{doi: #1}\else
  \providecommand{\doi}{doi: \begingroup \urlstyle{rm}\Url}\fi

\bibitem[Abnar \& Zuidema(2020)Abnar and Zuidema]{58}
Abnar, S. and Zuidema, W.~H.
\newblock Quantifying attention flow in transformers.
\newblock In Jurafsky, D., Chai, J., Schluter, N., and Tetreault, J.~R. (eds.),
  \emph{Proceedings of the 58th Annual Meeting of the Association for
  Computational Linguistics, {ACL} 2020, Online, July 5-10, 2020}, pp.\
  4190--4197. Association for Computational Linguistics, 2020.
\newblock \doi{10.18653/V1/2020.ACL-MAIN.385}.
\newblock URL \url{https://doi.org/10.18653/v1/2020.acl-main.385}.

\bibitem[Barkan et~al.(2023)Barkan, Elisha, Weill, Asher, Eshel, and
  Koenigstein]{47}
Barkan, O., Elisha, Y., Weill, J., Asher, Y., Eshel, A., and Koenigstein, N.
\newblock Deep integrated explanations.
\newblock In Frommholz, I., Hopfgartner, F., Lee, M., Oakes, M., Lalmas, M.,
  Zhang, M., and Santos, R. L.~T. (eds.), \emph{Proceedings of the 32nd {ACM}
  International Conference on Information and Knowledge Management, {CIKM}
  2023, Birmingham, United Kingdom, October 21-25, 2023}, pp.\  57--67. {ACM},
  2023.
\newblock \doi{10.1145/3583780.3614836}.
\newblock URL \url{https://doi.org/10.1145/3583780.3614836}.

\bibitem[Bastani et~al.(2017)Bastani, Kim, and
  Bastani]{DBLP:journals/corr/BastaniKB17a}
Bastani, O., Kim, C., and Bastani, H.
\newblock Interpretability via model extraction.
\newblock \emph{CoRR}, abs/1706.09773, 2017.
\newblock URL \url{http://arxiv.org/abs/1706.09773}.

\bibitem[Binder et~al.(2016)Binder, Montavon, Lapuschkin, M{\"{u}}ller, and
  Samek]{53}
Binder, A., Montavon, G., Lapuschkin, S., M{\"{u}}ller, K., and Samek, W.
\newblock Layer-wise relevance propagation for neural networks with local
  renormalization layers.
\newblock In Villa, A. E.~P., Masulli, P., and Rivero, A. J.~P. (eds.),
  \emph{Artificial Neural Networks and Machine Learning - {ICANN} 2016 - 25th
  International Conference on Artificial Neural Networks, Barcelona, Spain,
  September 6-9, 2016, Proceedings, Part {II}}, volume 9887 of \emph{Lecture
  Notes in Computer Science}, pp.\  63--71. Springer, 2016.
\newblock \doi{10.1007/978-3-319-44781-0\_8}.
\newblock URL \url{https://doi.org/10.1007/978-3-319-44781-0\_8}.

\bibitem[Changpinyo et~al.(2021)Changpinyo, Sharma, Ding, and Soricut]{add1_1}
Changpinyo, S., Sharma, P., Ding, N., and Soricut, R.
\newblock Conceptual 12m: Pushing web-scale image-text pre-training to
  recognize long-tail visual concepts.
\newblock In \emph{{IEEE} Conference on Computer Vision and Pattern
  Recognition, {CVPR} 2021, virtual, June 19-25, 2021}, pp.\  3558--3568.
  Computer Vision Foundation / {IEEE}, 2021.
\newblock \doi{10.1109/CVPR46437.2021.00356}.
\newblock URL
  \url{https://openaccess.thecvf.com/content/CVPR2021/html/Changpinyo\_Conceptual\_12M\_Pushing\_Web-Scale\_Image-Text\_Pre-Training\_To\_Recognize\_Long-Tail\_Visual\_CVPR\_2021\_paper.html}.

\bibitem[Chefer et~al.(2021{\natexlab{a}})Chefer, Gur, and Wolf]{48}
Chefer, H., Gur, S., and Wolf, L.
\newblock Generic attention-model explainability for interpreting bi-modal and
  encoder-decoder transformers.
\newblock In \emph{2021 {IEEE/CVF} International Conference on Computer Vision,
  {ICCV} 2021, Montreal, QC, Canada, October 10-17, 2021}, pp.\  387--396.
  {IEEE}, 2021{\natexlab{a}}.
\newblock \doi{10.1109/ICCV48922.2021.00045}.
\newblock URL \url{https://doi.org/10.1109/ICCV48922.2021.00045}.

\bibitem[Chefer et~al.(2021{\natexlab{b}})Chefer, Gur, and Wolf]{49}
Chefer, H., Gur, S., and Wolf, L.
\newblock Transformer interpretability beyond attention visualization.
\newblock In \emph{{IEEE} Conference on Computer Vision and Pattern
  Recognition, {CVPR} 2021, virtual, June 19-25, 2021}, pp.\  782--791.
  Computer Vision Foundation / {IEEE}, 2021{\natexlab{b}}.
\newblock \doi{10.1109/CVPR46437.2021.00084}.
\newblock URL
  \url{https://openaccess.thecvf.com/content/CVPR2021/html/Chefer\_Transformer\_Interpretability\_Beyond\_Attention\_Visualization\_CVPR\_2021\_paper.html}.

\bibitem[Chen et~al.(2023)Chen, Li, Yu, Dou, and Xiong]{46}
Chen, J., Li, X., Yu, L., Dou, D., and Xiong, H.
\newblock Beyond intuition: Rethinking token attributions inside transformers.
\newblock \emph{Trans. Mach. Learn. Res.}, 2023, 2023.
\newblock URL \url{https://openreview.net/forum?id=rm0zIzlhcX}.

\bibitem[Chen et~al.(2021)Chen, Wu, Wang, Liu, and Li]{12}
Chen, X., Wu, Y., Wang, Z., Liu, S., and Li, J.
\newblock Developing real-time streaming transformer transducer for speech
  recognition on large-scale dataset.
\newblock In \emph{{IEEE} International Conference on Acoustics, Speech and
  Signal Processing, {ICASSP} 2021, Toronto, ON, Canada, June 6-11, 2021}, pp.\
   5904--5908. {IEEE}, 2021.
\newblock \doi{10.1109/ICASSP39728.2021.9413535}.
\newblock URL \url{https://doi.org/10.1109/ICASSP39728.2021.9413535}.

\bibitem[Cui et~al.(2023)Cui, Yang, and Yao]{38}
Cui, Y., Yang, Z., and Yao, X.
\newblock Efficient and effective text encoding for chinese llama and alpaca.
\newblock \emph{CoRR}, abs/2304.08177, 2023.
\newblock \doi{10.48550/ARXIV.2304.08177}.
\newblock URL \url{https://doi.org/10.48550/arXiv.2304.08177}.

\bibitem[Dong et~al.(2018)Dong, Xu, and Xu]{13}
Dong, L., Xu, S., and Xu, B.
\newblock Speech-transformer: {A} no-recurrence sequence-to-sequence model for
  speech recognition.
\newblock In \emph{2018 {IEEE} International Conference on Acoustics, Speech
  and Signal Processing, {ICASSP} 2018, Calgary, AB, Canada, April 15-20,
  2018}, pp.\  5884--5888. {IEEE}, 2018.
\newblock \doi{10.1109/ICASSP.2018.8462506}.
\newblock URL \url{https://doi.org/10.1109/ICASSP.2018.8462506}.

\bibitem[Dosovitskiy et~al.(2021)Dosovitskiy, Beyer, Kolesnikov, Weissenborn,
  Zhai, Unterthiner, Dehghani, Minderer, Heigold, Gelly, Uszkoreit, and
  Houlsby]{6}
Dosovitskiy, A., Beyer, L., Kolesnikov, A., Weissenborn, D., Zhai, X.,
  Unterthiner, T., Dehghani, M., Minderer, M., Heigold, G., Gelly, S.,
  Uszkoreit, J., and Houlsby, N.
\newblock An image is worth 16x16 words: Transformers for image recognition at
  scale.
\newblock In \emph{9th International Conference on Learning Representations,
  {ICLR} 2021, Virtual Event, Austria, May 3-7, 2021}. OpenReview.net, 2021.
\newblock URL \url{https://openreview.net/forum?id=YicbFdNTTy}.

\bibitem[Englebert et~al.(2023)Englebert, Stassin, Nanfack, Mahmoudi, Siebert,
  Cornu, and Vleeschouwer]{DBLP:conf/iccvw/EnglebertSNMSCV23}
Englebert, A., Stassin, S., Nanfack, G., Mahmoudi, S.~A., Siebert, X., Cornu,
  O., and Vleeschouwer, C.~D.
\newblock Explaining through transformer input sampling.
\newblock In \emph{{IEEE/CVF} International Conference on Computer Vision,
  {ICCV} 2023 - Workshops, Paris, France, October 2-6, 2023}, pp.\  806--815.
  {IEEE}, 2023.
\newblock \doi{10.1109/ICCVW60793.2023.00088}.
\newblock URL \url{https://doi.org/10.1109/ICCVW60793.2023.00088}.

\bibitem[Hu et~al.(2020)Hu, Singh, Darrell, and Rohrbach]{21}
Hu, R., Singh, A., Darrell, T., and Rohrbach, M.
\newblock Iterative answer prediction with pointer-augmented multimodal
  transformers for textvqa.
\newblock In \emph{2020 {IEEE/CVF} Conference on Computer Vision and Pattern
  Recognition, {CVPR} 2020, Seattle, WA, USA, June 13-19, 2020}, pp.\
  9989--9999. Computer Vision Foundation / {IEEE}, 2020.
\newblock \doi{10.1109/CVPR42600.2020.01001}.
\newblock URL
  \url{https://openaccess.thecvf.com/content\_CVPR\_2020/html/Hu\_Iterative\_Answer\_Prediction\_With\_Pointer-Augmented\_Multimodal\_Transformers\_for\_TextVQA\_CVPR\_2020\_paper.html}.

\bibitem[Ji et~al.(2021)Ji, Zhang, Wang, Li, Wu, Zhang, and Luo]{29}
Ji, Y., Zhang, R., Wang, H., Li, Z., Wu, L., Zhang, S., and Luo, P.
\newblock Multi-compound transformer for accurate biomedical image
  segmentation.
\newblock In de~Bruijne, M., Cattin, P.~C., Cotin, S., Padoy, N., Speidel, S.,
  Zheng, Y., and Essert, C. (eds.), \emph{Medical Image Computing and Computer
  Assisted Intervention - {MICCAI} 2021 - 24th International Conference,
  Strasbourg, France, September 27 - October 1, 2021, Proceedings, Part {I}},
  volume 12901 of \emph{Lecture Notes in Computer Science}, pp.\  326--336.
  Springer, 2021.
\newblock \doi{10.1007/978-3-030-87193-2\_31}.
\newblock URL \url{https://doi.org/10.1007/978-3-030-87193-2\_31}.

\bibitem[Jiang et~al.(2021)Jiang, Zhang, Hou, Cheng, and Wei]{52}
Jiang, P., Zhang, C., Hou, Q., Cheng, M., and Wei, Y.
\newblock Layercam: Exploring hierarchical class activation maps for
  localization.
\newblock \emph{{IEEE} Trans. Image Process.}, 30:\penalty0 5875--5888, 2021.
\newblock \doi{10.1109/TIP.2021.3089943}.
\newblock URL \url{https://doi.org/10.1109/TIP.2021.3089943}.

\bibitem[Li et~al.(2022)Li, Li, Xiong, and Hoi]{add4_1}
Li, J., Li, D., Xiong, C., and Hoi, S. C.~H.
\newblock {BLIP:} bootstrapping language-image pre-training for unified
  vision-language understanding and generation.
\newblock In Chaudhuri, K., Jegelka, S., Song, L., Szepesv{\'{a}}ri, C., Niu,
  G., and Sabato, S. (eds.), \emph{International Conference on Machine
  Learning, {ICML} 2022, 17-23 July 2022, Baltimore, Maryland, {USA}}, volume
  162 of \emph{Proceedings of Machine Learning Research}, pp.\  12888--12900.
  {PMLR}, 2022.
\newblock URL \url{https://proceedings.mlr.press/v162/li22n.html}.

\bibitem[Li et~al.(2019)Li, Yatskar, Yin, Hsieh, and Chang]{22}
Li, L.~H., Yatskar, M., Yin, D., Hsieh, C., and Chang, K.
\newblock Visualbert: {A} simple and performant baseline for vision and
  language.
\newblock \emph{CoRR}, abs/1908.03557, 2019.
\newblock URL \url{http://arxiv.org/abs/1908.03557}.

\bibitem[Li et~al.(2021)Li, Gao, Niu, Xiao, Liu, Liu, Wu, and Wang]{23}
Li, W., Gao, C., Niu, G., Xiao, X., Liu, H., Liu, J., Wu, H., and Wang, H.
\newblock {UNIMO:} towards unified-modal understanding and generation via
  cross-modal contrastive learning.
\newblock In Zong, C., Xia, F., Li, W., and Navigli, R. (eds.),
  \emph{Proceedings of the 59th Annual Meeting of the Association for
  Computational Linguistics and the 11th International Joint Conference on
  Natural Language Processing, {ACL/IJCNLP} 2021, (Volume 1: Long Papers),
  Virtual Event, August 1-6, 2021}, pp.\  2592--2607. Association for
  Computational Linguistics, 2021.
\newblock \doi{10.18653/V1/2021.ACL-LONG.202}.
\newblock URL \url{https://doi.org/10.18653/v1/2021.acl-long.202}.

\bibitem[Lin et~al.(2014)Lin, Maire, Belongie, Hays, Perona, Ramanan,
  Doll{\'{a}}r, and Zitnick]{coco}
Lin, T., Maire, M., Belongie, S.~J., Hays, J., Perona, P., Ramanan, D.,
  Doll{\'{a}}r, P., and Zitnick, C.~L.
\newblock Microsoft {COCO:} common objects in context.
\newblock In Fleet, D.~J., Pajdla, T., Schiele, B., and Tuytelaars, T. (eds.),
  \emph{Computer Vision - {ECCV} 2014 - 13th European Conference, Zurich,
  Switzerland, September 6-12, 2014, Proceedings, Part {V}}, volume 8693 of
  \emph{Lecture Notes in Computer Science}, pp.\  740--755. Springer, 2014.
\newblock \doi{10.1007/978-3-319-10602-1\_48}.
\newblock URL \url{https://doi.org/10.1007/978-3-319-10602-1\_48}.

\bibitem[Liu et~al.(2021)Liu, Lin, Cao, Hu, Wei, Zhang, Lin, and Guo]{7}
Liu, Z., Lin, Y., Cao, Y., Hu, H., Wei, Y., Zhang, Z., Lin, S., and Guo, B.
\newblock Swin transformer: Hierarchical vision transformer using shifted
  windows.
\newblock In \emph{2021 {IEEE/CVF} International Conference on Computer Vision,
  {ICCV} 2021, Montreal, QC, Canada, October 10-17, 2021}, pp.\  9992--10002.
  {IEEE}, 2021.
\newblock \doi{10.1109/ICCV48922.2021.00986}.
\newblock URL \url{https://doi.org/10.1109/ICCV48922.2021.00986}.

\bibitem[Luo et~al.(2022)Luo, Ji, Zhong, Chen, Lei, Duan, and Li]{add2_1}
Luo, H., Ji, L., Zhong, M., Chen, Y., Lei, W., Duan, N., and Li, T.
\newblock Clip4clip: An empirical study of {CLIP} for end to end video clip
  retrieval and captioning.
\newblock \emph{Neurocomputing}, 508:\penalty0 293--304, 2022.
\newblock \doi{10.1016/J.NEUCOM.2022.07.028}.
\newblock URL \url{https://doi.org/10.1016/j.neucom.2022.07.028}.

\bibitem[Mariotti et~al.(2023)Mariotti, Sivaprasad, and
  Alonso{-}Moral]{DBLP:conf/xai/MariottiSA23}
Mariotti, E., Sivaprasad, A., and Alonso{-}Moral, J.~M.
\newblock Beyond prediction similarity: Shapgap for evaluating faithful
  surrogate models in {XAI}.
\newblock In Longo, L. (ed.), \emph{Explainable Artificial Intelligence - First
  World Conference, xAI 2023, Lisbon, Portugal, July 26-28, 2023, Proceedings,
  Part {I}}, Communications in Computer and Information Science, pp.\
  160--173. Springer, 2023.
\newblock \doi{10.1007/978-3-031-44064-9\_10}.
\newblock URL \url{https://doi.org/10.1007/978-3-031-44064-9\_10}.

\bibitem[Michel et~al.(2019)Michel, Levy, and Neubig]{57}
Michel, P., Levy, O., and Neubig, G.
\newblock Are sixteen heads really better than one?
\newblock In Wallach, H.~M., Larochelle, H., Beygelzimer, A.,
  d'Alch{\'{e}}{-}Buc, F., Fox, E.~B., and Garnett, R. (eds.), \emph{Advances
  in Neural Information Processing Systems 32: Annual Conference on Neural
  Information Processing Systems 2019, NeurIPS 2019, December 8-14, 2019,
  Vancouver, BC, Canada}, pp.\  14014--14024, 2019.
\newblock URL
  \url{https://proceedings.neurips.cc/paper/2019/hash/2c601ad9d2ff9bc8b282670cdd54f69f-Abstract.html}.

\bibitem[Radford et~al.(2021)Radford, Kim, Hallacy, Ramesh, Goh, Agarwal,
  Sastry, Askell, Mishkin, Clark, Krueger, and Sutskever]{CLIP}
Radford, A., Kim, J.~W., Hallacy, C., Ramesh, A., Goh, G., Agarwal, S., Sastry,
  G., Askell, A., Mishkin, P., Clark, J., Krueger, G., and Sutskever, I.
\newblock Learning transferable visual models from natural language
  supervision.
\newblock In Meila, M. and Zhang, T. (eds.), \emph{Proceedings of the 38th
  International Conference on Machine Learning, {ICML} 2021, 18-24 July 2021,
  Virtual Event}, volume 139 of \emph{Proceedings of Machine Learning
  Research}, pp.\  8748--8763. {PMLR}, 2021.
\newblock URL \url{http://proceedings.mlr.press/v139/radford21a.html}.

\bibitem[Recht et~al.(2019)Recht, Roelofs, Schmidt, and Shankar]{newimagenet}
Recht, B., Roelofs, R., Schmidt, L., and Shankar, V.
\newblock Do imagenet classifiers generalize to imagenet?
\newblock \emph{CoRR}, abs/1902.10811, 2019.
\newblock URL \url{http://arxiv.org/abs/1902.10811}.

\bibitem[Ribeiro et~al.(2016)Ribeiro, Singh, and
  Guestrin]{DBLP:conf/kdd/Ribeiro0G16}
Ribeiro, M.~T., Singh, S., and Guestrin, C.
\newblock "why should {I} trust you?": Explaining the predictions of any
  classifier.
\newblock In Krishnapuram, B., Shah, M., Smola, A.~J., Aggarwal, C.~C., Shen,
  D., and Rastogi, R. (eds.), \emph{Proceedings of the 22nd {ACM} {SIGKDD}
  International Conference on Knowledge Discovery and Data Mining, San
  Francisco, CA, USA, August 13-17, 2016}, pp.\  1135--1144. {ACM}, 2016.
\newblock \doi{10.1145/2939672.2939778}.
\newblock URL \url{https://doi.org/10.1145/2939672.2939778}.

\bibitem[Russakovsky et~al.(2015)Russakovsky, Deng, Su, Krause, Satheesh, Ma,
  Huang, Karpathy, Khosla, Bernstein, Berg, and Fei{-}Fei]{62}
Russakovsky, O., Deng, J., Su, H., Krause, J., Satheesh, S., Ma, S., Huang, Z.,
  Karpathy, A., Khosla, A., Bernstein, M.~S., Berg, A.~C., and Fei{-}Fei, L.
\newblock Imagenet large scale visual recognition challenge.
\newblock \emph{Int. J. Comput. Vis.}, 115\penalty0 (3):\penalty0 211--252,
  2015.
\newblock \doi{10.1007/S11263-015-0816-Y}.
\newblock URL \url{https://doi.org/10.1007/s11263-015-0816-y}.

\bibitem[Selvaraju et~al.(2017)Selvaraju, Cogswell, Das, Vedantam, Parikh, and
  Batra]{51}
Selvaraju, R.~R., Cogswell, M., Das, A., Vedantam, R., Parikh, D., and Batra,
  D.
\newblock Grad-cam: Visual explanations from deep networks via gradient-based
  localization.
\newblock In \emph{{IEEE} International Conference on Computer Vision, {ICCV}
  2017, Venice, Italy, October 22-29, 2017}, pp.\  618--626. {IEEE} Computer
  Society, 2017.
\newblock \doi{10.1109/ICCV.2017.74}.
\newblock URL \url{https://doi.org/10.1109/ICCV.2017.74}.

\bibitem[Thiagarajan et~al.(2016)Thiagarajan, Kailkhura, Sattigeri, and
  Ramamurthy]{DBLP:journals/corr/ThiagarajanKSR16}
Thiagarajan, J.~J., Kailkhura, B., Sattigeri, P., and Ramamurthy, K.~N.
\newblock Treeview: Peeking into deep neural networks via feature-space
  partitioning.
\newblock \emph{CoRR}, abs/1611.07429, 2016.
\newblock URL \url{http://arxiv.org/abs/1611.07429}.

\bibitem[Touvron et~al.(2023{\natexlab{a}})Touvron, Lavril, Izacard, Martinet,
  Lachaux, Lacroix, Rozi{\`{e}}re, Goyal, Hambro, Azhar, Rodriguez, Joulin,
  Grave, and Lample]{41}
Touvron, H., Lavril, T., Izacard, G., Martinet, X., Lachaux, M., Lacroix, T.,
  Rozi{\`{e}}re, B., Goyal, N., Hambro, E., Azhar, F., Rodriguez, A., Joulin,
  A., Grave, E., and Lample, G.
\newblock Llama: Open and efficient foundation language models.
\newblock \emph{CoRR}, abs/2302.13971, 2023{\natexlab{a}}.
\newblock \doi{10.48550/ARXIV.2302.13971}.
\newblock URL \url{https://doi.org/10.48550/arXiv.2302.13971}.

\bibitem[Touvron et~al.(2023{\natexlab{b}})Touvron, Martin, Stone, Albert,
  Almahairi, Babaei, Bashlykov, Batra, Bhargava, Bhosale, Bikel, Blecher,
  Canton{-}Ferrer, Chen, Cucurull, Esiobu, Fernandes, Fu, Fu, Fuller, Gao,
  Goswami, Goyal, Hartshorn, Hosseini, Hou, Inan, Kardas, Kerkez, Khabsa,
  Kloumann, Korenev, Koura, Lachaux, Lavril, Lee, Liskovich, Lu, Mao, Martinet,
  Mihaylov, Mishra, Molybog, Nie, Poulton, Reizenstein, Rungta, Saladi,
  Schelten, Silva, Smith, Subramanian, Tan, Tang, Taylor, Williams, Kuan, Xu,
  Yan, Zarov, Zhang, Fan, Kambadur, Narang, Rodriguez, Stojnic, Edunov, and
  Scialom]{42}
Touvron, H., Martin, L., Stone, K., Albert, P., Almahairi, A., Babaei, Y.,
  Bashlykov, N., Batra, S., Bhargava, P., Bhosale, S., Bikel, D., Blecher, L.,
  Canton{-}Ferrer, C., Chen, M., Cucurull, G., Esiobu, D., Fernandes, J., Fu,
  J., Fu, W., Fuller, B., Gao, C., Goswami, V., Goyal, N., Hartshorn, A.,
  Hosseini, S., Hou, R., Inan, H., Kardas, M., Kerkez, V., Khabsa, M.,
  Kloumann, I., Korenev, A., Koura, P.~S., Lachaux, M., Lavril, T., Lee, J.,
  Liskovich, D., Lu, Y., Mao, Y., Martinet, X., Mihaylov, T., Mishra, P.,
  Molybog, I., Nie, Y., Poulton, A., Reizenstein, J., Rungta, R., Saladi, K.,
  Schelten, A., Silva, R., Smith, E.~M., Subramanian, R., Tan, X.~E., Tang, B.,
  Taylor, R., Williams, A., Kuan, J.~X., Xu, P., Yan, Z., Zarov, I., Zhang, Y.,
  Fan, A., Kambadur, M., Narang, S., Rodriguez, A., Stojnic, R., Edunov, S.,
  and Scialom, T.
\newblock Llama 2: Open foundation and fine-tuned chat models.
\newblock \emph{CoRR}, abs/2307.09288, 2023{\natexlab{b}}.
\newblock \doi{10.48550/ARXIV.2307.09288}.
\newblock URL \url{https://doi.org/10.48550/arXiv.2307.09288}.

\bibitem[Vaswani et~al.(2017)Vaswani, Shazeer, Parmar, Uszkoreit, Jones, Gomez,
  Kaiser, and Polosukhin]{1}
Vaswani, A., Shazeer, N., Parmar, N., Uszkoreit, J., Jones, L., Gomez, A.~N.,
  Kaiser, L., and Polosukhin, I.
\newblock Attention is all you need.
\newblock In Guyon, I., von Luxburg, U., Bengio, S., Wallach, H.~M., Fergus,
  R., Vishwanathan, S. V.~N., and Garnett, R. (eds.), \emph{Advances in Neural
  Information Processing Systems 30: Annual Conference on Neural Information
  Processing Systems 2017, December 4-9, 2017, Long Beach, CA, {USA}}, pp.\
  5998--6008, 2017.
\newblock URL
  \url{https://proceedings.neurips.cc/paper/2017/hash/3f5ee243547dee91fbd053c1c4a845aa-Abstract.html}.

\bibitem[Wang et~al.(2021)Wang, Lai, and Kong]{30}
Wang, T., Lai, Z., and Kong, H.
\newblock Tfnet: Transformer fusion network for ultrasound image segmentation.
\newblock In Wallraven, C., Liu, Q., and Nagahara, H. (eds.), \emph{Pattern
  Recognition - 6th Asian Conference, {ACPR} 2021, Jeju Island, South Korea,
  November 9-12, 2021, Revised Selected Papers, Part {I}}, volume 13188 of
  \emph{Lecture Notes in Computer Science}, pp.\  314--325. Springer, 2021.
\newblock \doi{10.1007/978-3-031-02375-0\_23}.
\newblock URL \url{https://doi.org/10.1007/978-3-031-02375-0\_23}.

\bibitem[Wang et~al.(2022)Wang, Lu, Li, Tao, Guo, Gong, and Liu]{add3_1}
Wang, Z., Lu, Y., Li, Q., Tao, X., Guo, Y., Gong, M., and Liu, T.
\newblock {CRIS:} clip-driven referring image segmentation.
\newblock In \emph{{IEEE/CVF} Conference on Computer Vision and Pattern
  Recognition, {CVPR} 2022, New Orleans, LA, USA, June 18-24, 2022}, pp.\
  11676--11685. {IEEE}, 2022.
\newblock \doi{10.1109/CVPR52688.2022.01139}.
\newblock URL \url{https://doi.org/10.1109/CVPR52688.2022.01139}.

\bibitem[Xie et~al.(2023)Xie, Li, Cao, and Zhang]{DBLP:conf/ijcai/Xie0CZ23}
Xie, W., Li, X., Cao, C.~C., and Zhang, N.~L.
\newblock Vit-cx: Causal explanation of vision transformers.
\newblock In \emph{Proceedings of the Thirty-Second International Joint
  Conference on Artificial Intelligence, {IJCAI} 2023, 19th-25th August 2023,
  Macao, SAR, China}, pp.\  1569--1577. ijcai.org, 2023.
\newblock \doi{10.24963/IJCAI.2023/174}.
\newblock URL \url{https://doi.org/10.24963/ijcai.2023/174}.

\bibitem[Xie et~al.(2021)Xie, Zhang, Shen, and Xia]{28}
Xie, Y., Zhang, J., Shen, C., and Xia, Y.
\newblock Cotr: Efficiently bridging {CNN} and transformer for 3d medical image
  segmentation.
\newblock In de~Bruijne, M., Cattin, P.~C., Cotin, S., Padoy, N., Speidel, S.,
  Zheng, Y., and Essert, C. (eds.), \emph{Medical Image Computing and Computer
  Assisted Intervention - {MICCAI} 2021 - 24th International Conference,
  Strasbourg, France, September 27 - October 1, 2021, Proceedings, Part {III}},
  volume 12903 of \emph{Lecture Notes in Computer Science}, pp.\  171--180.
  Springer, 2021.
\newblock \doi{10.1007/978-3-030-87199-4\_16}.
\newblock URL \url{https://doi.org/10.1007/978-3-030-87199-4\_16}.

\bibitem[Xu et~al.(2022)Xu, Mello, Liu, Byeon, Breuel, Kautz, and Wang]{add3_2}
Xu, J., Mello, S.~D., Liu, S., Byeon, W., Breuel, T.~M., Kautz, J., and Wang,
  X.
\newblock Groupvit: Semantic segmentation emerges from text supervision.
\newblock In \emph{{IEEE/CVF} Conference on Computer Vision and Pattern
  Recognition, {CVPR} 2022, New Orleans, LA, USA, June 18-24, 2022}, pp.\
  18113--18123. {IEEE}, 2022.
\newblock \doi{10.1109/CVPR52688.2022.01760}.
\newblock URL \url{https://doi.org/10.1109/CVPR52688.2022.01760}.

\bibitem[Yuan et~al.(2021)Yuan, Li, Xiong, Cao, and Dou]{yuan2021explaining}
Yuan, T., Li, X., Xiong, H., Cao, H., and Dou, D.
\newblock Explaining information flow inside vision transformers using markov
  chain.
\newblock In \emph{eXplainable AI approaches for debugging and diagnosis.},
  2021.
\newblock URL \url{https://openreview.net/forum?id=TT-cf6QSDaQ}.

\bibitem[Zhao et~al.(2024)Zhao, Wang, Zeng, Zhao, and Chan]{gradeclip}
Zhao, C., Wang, K., Zeng, X., Zhao, R., and Chan, A.~B.
\newblock Gradient-based visual explanation for transformer-based {CLIP}.
\newblock In \emph{Forty-first International Conference on Machine Learning,
  {ICML} 2024, Vienna, Austria, July 21-27, 2024}. OpenReview.net, 2024.
\newblock URL \url{https://openreview.net/forum?id=WT4X3QYopC}.

\bibitem[Zhong et~al.(2022)Zhong, Yang, Zhang, Li, Codella, Li, Zhou, Dai,
  Yuan, Li, and Gao]{add6_1}
Zhong, Y., Yang, J., Zhang, P., Li, C., Codella, N., Li, L.~H., Zhou, L., Dai,
  X., Yuan, L., Li, Y., and Gao, J.
\newblock Regionclip: Region-based language-image pretraining.
\newblock In \emph{{IEEE/CVF} Conference on Computer Vision and Pattern
  Recognition, {CVPR} 2022, New Orleans, LA, USA, June 18-24, 2022}, pp.\
  16772--16782. {IEEE}, 2022.
\newblock \doi{10.1109/CVPR52688.2022.01629}.
\newblock URL \url{https://doi.org/10.1109/CVPR52688.2022.01629}.

\bibitem[Zhou et~al.(2016)Zhou, Khosla, Lapedriza, Oliva, and Torralba]{50}
Zhou, B., Khosla, A., Lapedriza, {\`{A}}., Oliva, A., and Torralba, A.
\newblock Learning deep features for discriminative localization.
\newblock In \emph{2016 {IEEE} Conference on Computer Vision and Pattern
  Recognition, {CVPR} 2016, Las Vegas, NV, USA, June 27-30, 2016}, pp.\
  2921--2929. {IEEE} Computer Society, 2016.
\newblock \doi{10.1109/CVPR.2016.319}.
\newblock URL \url{https://doi.org/10.1109/CVPR.2016.319}.

\bibitem[Zhou et~al.(2022)Zhou, Yang, Loy, and Liu]{add5_1}
Zhou, K., Yang, J., Loy, C.~C., and Liu, Z.
\newblock Learning to prompt for vision-language models.
\newblock \emph{Int. J. Comput. Vis.}, 130\penalty0 (9):\penalty0 2337--2348,
  2022.
\newblock \doi{10.1007/S11263-022-01653-1}.
\newblock URL \url{https://doi.org/10.1007/s11263-022-01653-1}.

\end{thebibliography}
\bibliographystyle{icml2026}

\newpage
\appendix
\onecolumn

\section{Some Results of Subsection \ref{sec:ex veri}}\label{buchongdingliang}
\FloatBarrier

The experimental design is presented in Subsection \ref{sec:ex veri}. The minor discrepancies stem from slight differences in algorithm implementation or numerical calculation, floating point error, or random processes.

\begin{table*}[h]
    \centering
    \caption{Image interpretation experiments of Grad-ECLIP and Attention-ECLIP.}
    \label{table2}
    \begin{tabular}{@{}lrrrrrrrrrrr@{}}
    \toprule
    \multicolumn{12}{c}{Deletion}                                                                                                                                                                                                                                                                                                                                                                      \\ \midrule
                                                                     & \multicolumn{1}{c}{Original} & \multicolumn{1}{c}{Step 1} & \multicolumn{1}{c}{Step 2} & \multicolumn{1}{c}{Step 3} & \multicolumn{1}{c}{Step 4} & \multicolumn{1}{c}{Step 5} & \multicolumn{1}{c}{Step 6} & \multicolumn{1}{c}{Step 7} & \multicolumn{1}{c}{Step 8} & \multicolumn{1}{c}{Step 9} & \multicolumn{1}{c}{Step 10} \\ \midrule
    \begin{tabular}[c]{@{}l@{}}Grad-ECLIP\\ w/o $w_i$\end{tabular}      & 0.710                         & 0.491                      & 0.383                      & 0.298                      & 0.239                      & 0.183                      & 0.150                       & 0.118                      & 0.086                      & 0.077                      & 0.061                       \\
    \begin{tabular}[c]{@{}l@{}}Attention-ECLIP\\ w/o $w_i$\end{tabular} & 0.710                         & 0.497                      & 0.383                      & 0.299                      & 0.240                       & 0.182                      & 0.153                      & 0.113                      & 0.089                      & 0.078                      & 0.062                       \\
    Grad-ECLIP                                                       & 0.709                        & 0.516                      & 0.411                      & 0.311                      & 0.263                      & 0.216                      & 0.167                      & 0.132                      & 0.109                      & 0.083                      & 0.075                       \\
    Attention-ECLIP                                                  & 0.709                        & 0.514                      & 0.409                      & 0.310                       & 0.265                      & 0.217                      & 0.165                      & 0.129                      & 0.108                      & 0.083                      & 0.074                       \\ \midrule
    \multicolumn{12}{c}{Insertion}                                                                                                                                                                                                                                                                                                                                                                     \\ \midrule
                                                                     & \multicolumn{1}{c}{Original} & \multicolumn{1}{c}{Step 1} & \multicolumn{1}{c}{Step 2} & \multicolumn{1}{c}{Step 3} & \multicolumn{1}{c}{Step 4} & \multicolumn{1}{c}{Step 5} & \multicolumn{1}{c}{Step 6} & \multicolumn{1}{c}{Step 7} & \multicolumn{1}{c}{Step 8} & \multicolumn{1}{c}{Step 9} & \multicolumn{1}{c}{Step 10} \\ \midrule
    \begin{tabular}[c]{@{}l@{}}Grad-ECLIP\\ w/o $w_i$\end{tabular}      & 0.710                         & 0.083                      & 0.197                      & 0.284                      & 0.368                      & 0.457                      & 0.503                      & 0.539                      & 0.583                      & 0.598                      & 0.619                       \\
    \begin{tabular}[c]{@{}l@{}}Attention-ECLIP\\ w/o $w_i$\end{tabular} & 0.710                         & 0.083                      & 0.199                      & 0.284                      & 0.369                      & 0.453                      & 0.502                      & 0.541                      & 0.580                       & 0.597                      & 0.622                       \\
    Grad-ECLIP                                                       & 0.709                        & 0.066                      & 0.144                      & 0.258                      & 0.331                      & 0.397                      & 0.459                      & 0.501                      & 0.541                      & 0.575                      & 0.604                       \\
    Attention-ECLIP                                                  & 0.709                        & 0.066                      & 0.139                      & 0.255                      & 0.337                      & 0.398                      & 0.461                      & 0.507                      & 0.537                      & 0.571                      & 0.606                       \\ \bottomrule
    \end{tabular}
    \end{table*}

\begin{table*}[h]
    \centering
    \caption{Text interpretation experiments of Grad-ECLIP and Attention-ECLIP.}
    \label{table3}
    \begin{tabular}{@{}lllllll@{}}
    \toprule
                           & Original  & Step 1    & Step 2    & Step 3    & Step 4    & Step 5    \\ \midrule
    Grad-ECLIP w/o $w_i$      & 30.668344 & 27.860970    & 25.686476 & 24.065400 & 22.876960 & 21.950430 \\
    Attention-ECLIP w/o $w_i$ & 30.668344 & 27.860970    & 25.687056 & 24.065400 & 22.876960 & 21.950820 \\
    Grad-ECLIP             & 30.668344 & 27.769672 & 25.271976 & 23.324562 & 21.951883 & 21.137398    \\
    Attention-ECLIP        & 30.668344 & 27.769672  & 25.271976 & 23.324562  & 21.951883 & 21.136961  \\ \bottomrule
    \end{tabular}
    \vspace{-0.5cm}

    \end{table*}
\FloatBarrier

\section{Performance Comparison between the Original CLIP and the Modified CLIP}\label{comp}
\FloatBarrier

\begin{figure*}[h]
    \centering
    \includegraphics[width=0.7\columnwidth]{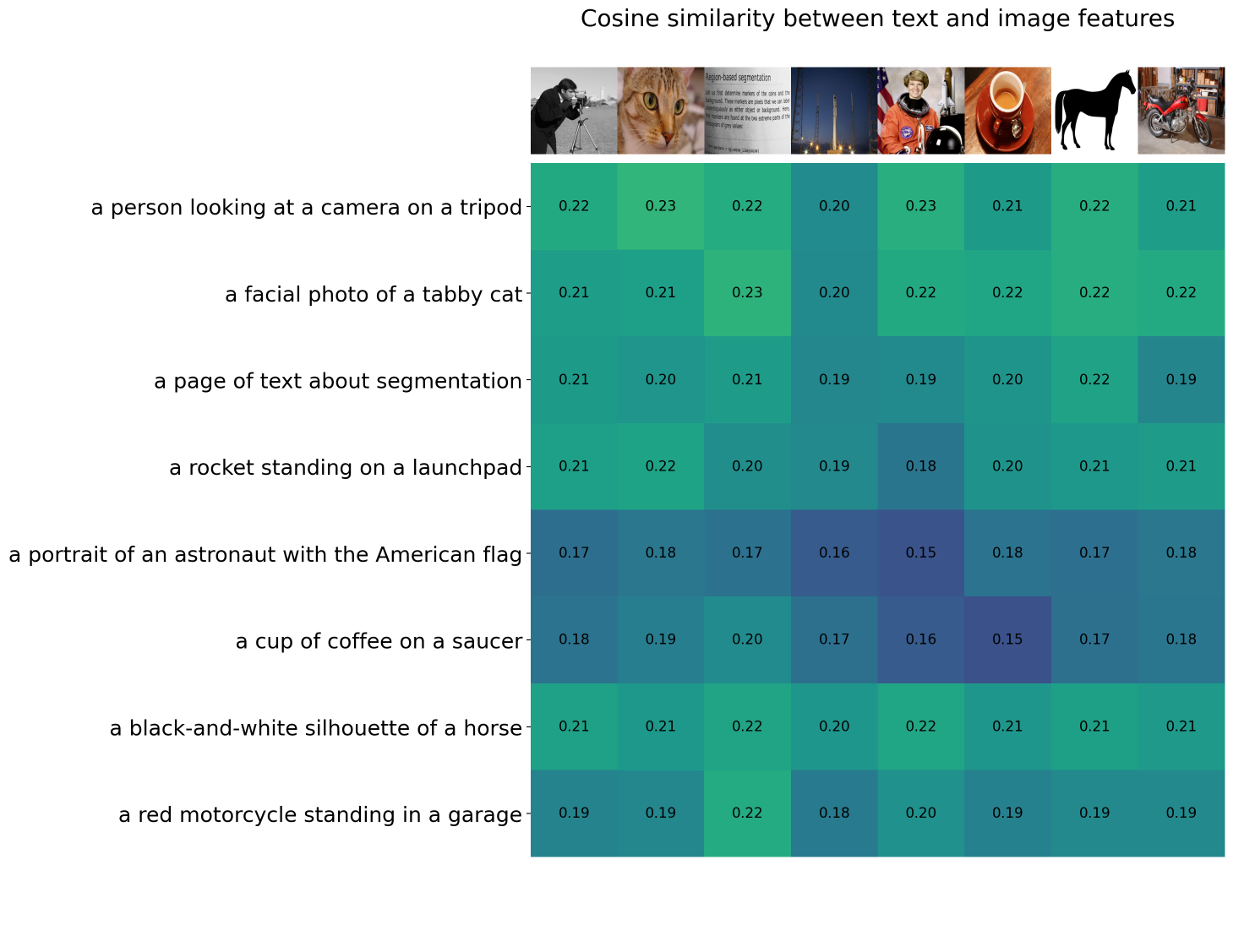} 
    \caption{Performance of the modified CLIP (the last attention layer of the image encoder is modified to single-head attention).}
    \label{clipzhao}
    \end{figure*}

\begin{figure*}[h]
    \centering
    \includegraphics[width=0.7\columnwidth]{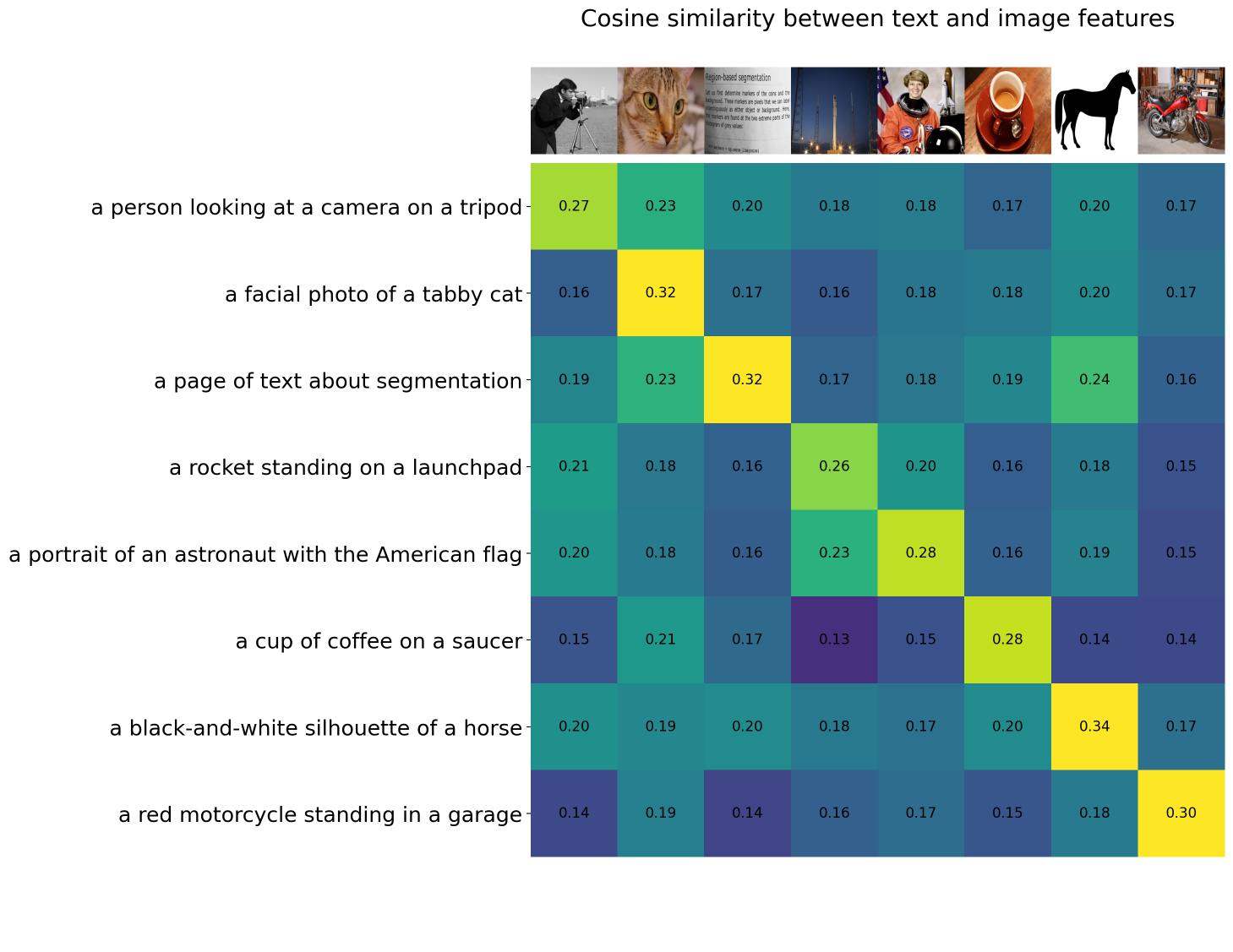} 
    \caption{Performance of the modified CLIP (the last attention layer of the text encoder is modified to single-head attention).}
    \label{clipzhao21}
    \vspace{-0.7cm}
    \end{figure*}

\begin{figure*}[h]
    \centering
    \includegraphics[width=0.7\columnwidth]{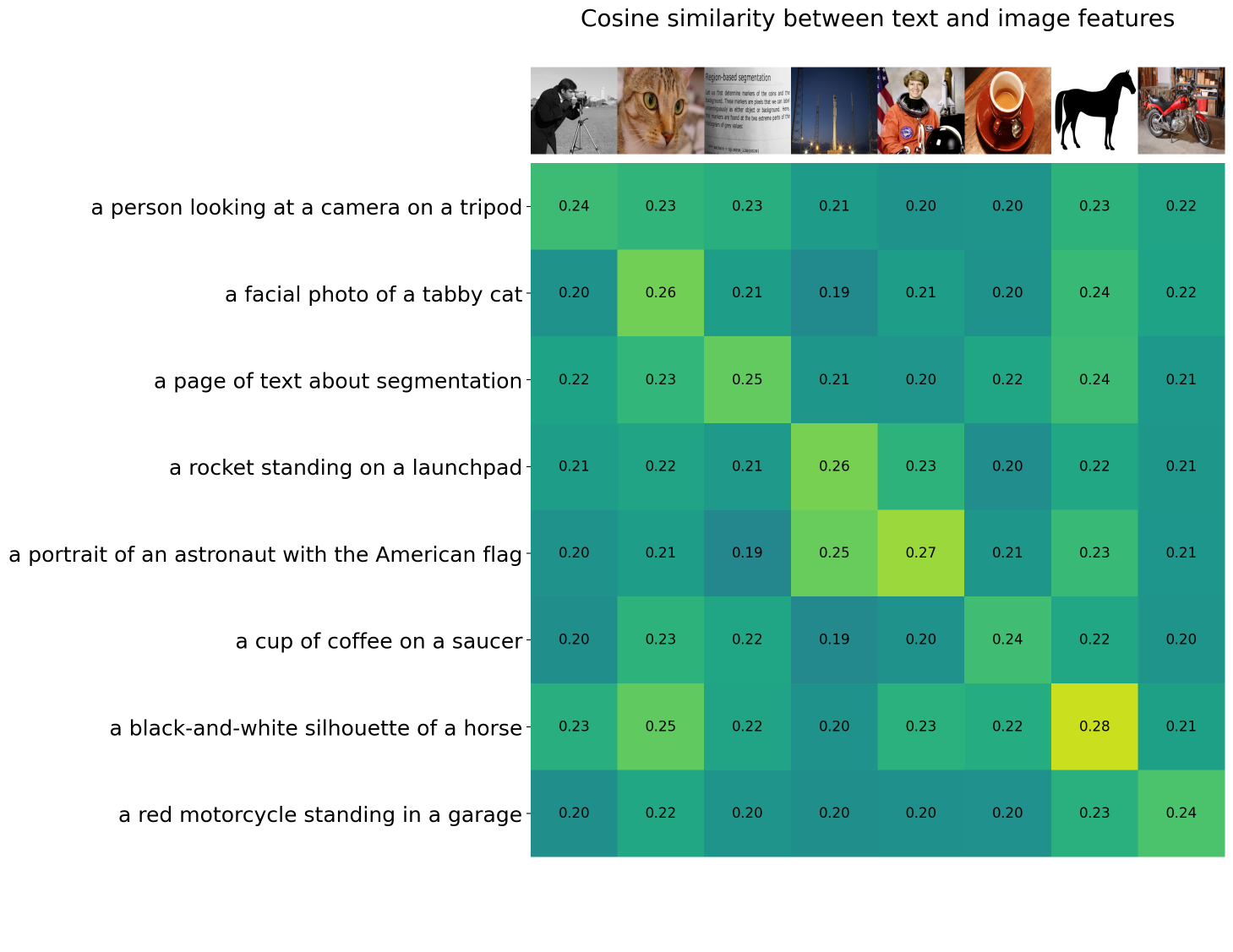}
    \caption{Performance of the modified CLIP (the last eight attention layers of the text encoder is modified to single-head attention).}
    \label{clipzhao28}
    \vspace{-0.7cm}
    \end{figure*}

\begin{figure*}[h]
    \centering
    \includegraphics[width=0.7\columnwidth]{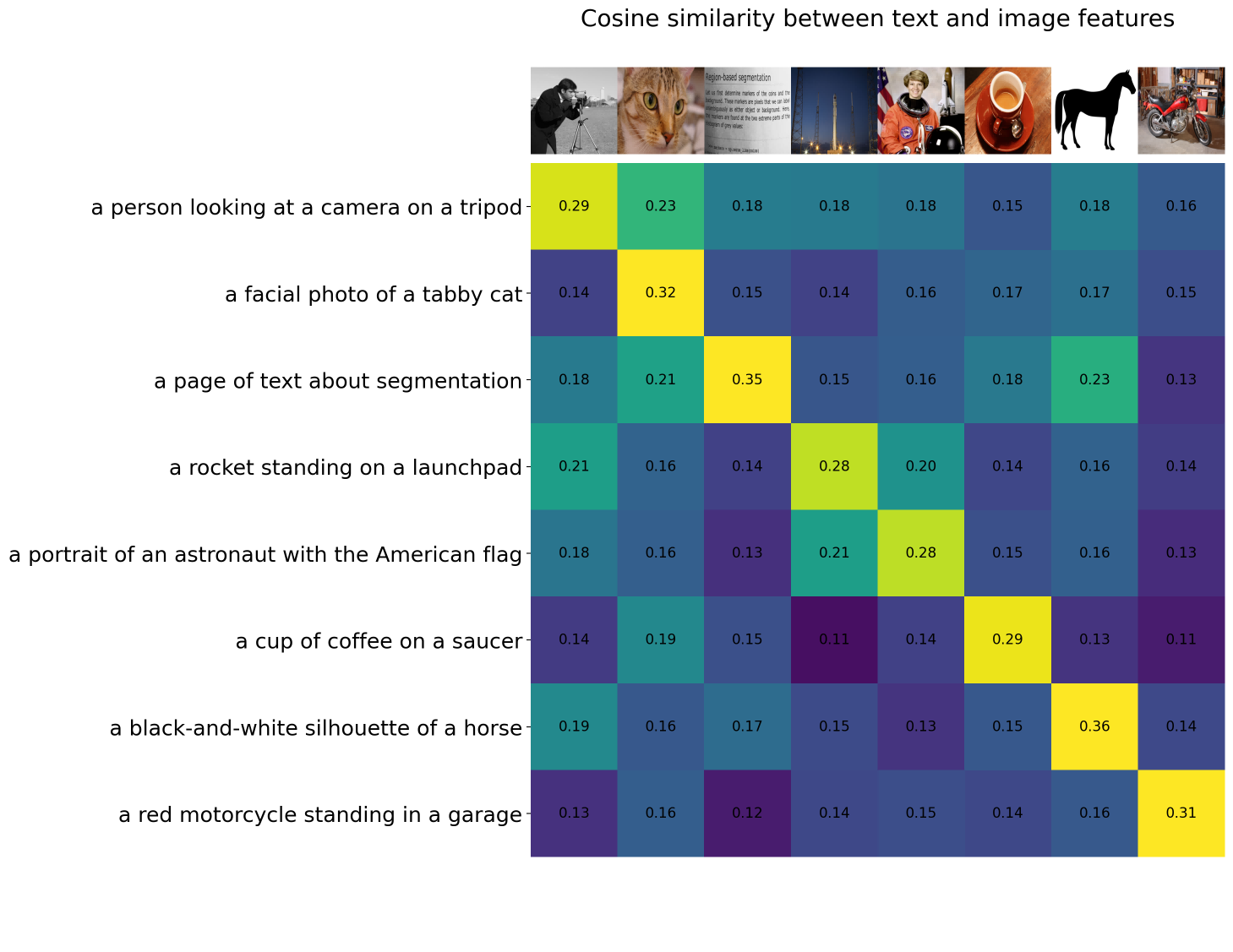}
    \caption{Performance of the original CLIP.}
    \label{clipyuan}
    \vspace{-0.7cm}
    \end{figure*}

Modifying the multi-head attention of one layer in the original model to single-head attention directly disrupts the model structure. It is almost inevitable that this will lead to a decline in model performance, and we believe there is hardly any need for verification. For the sake of rigor, we repeated the example experiment of CLIP (https://github.com/OpenAI/CLIP). 
We take the performance of the original CLIP as baseline (Figure \ref{clipyuan}).
As shown in Figure \ref{clipzhao}, the modified model (with the last attention layer of the image encoder is modified to single-head attention) is completely unable to correctly judge the relevance between images and text; performance of the modified model with text encoder modified is shown in Figure \ref{clipzhao21} and \ref{clipzhao28}, both the performance of the two modified model reduced, especially the last eight layer modified CLIP, as shown in Figure \ref{clipzhao28}.

Additionally, to test the model's performance on a broader scale, we replicated another experiment from the CLIP paper \cite{CLIP}, conducting a zero-shot capability test of the model using 10,000 samples from ImageNet-V2 \cite{newimagenet}. The experimental results (accuracy of the model in correctly predicting sample category) are shown in Table \ref{zeroshot}.

\begin{table}[]
    \caption{Zero-shot capability test.}
    \label{zeroshot}
    \begin{tabular}{@{}lllll@{}}
    \toprule
                          & Original CLIP & \begin{tabular}[c]{@{}l@{}}Modified CLIP\\ (the last layer of \\ image encoder modified)\end{tabular} & \begin{tabular}[c]{@{}l@{}}Modified CLIP\\ (the last layer of \\ text encoder modified)\end{tabular} & \begin{tabular}[c]{@{}l@{}}Modified CLIP\\ (the last eight layers of \\ text encoder modified)\end{tabular} \\ \midrule
    Top-1   accuracy (\%) & 55.90          & 2.22                                                                                                  & 51.93                                                                                                & 31.29                                                                                                       \\
    Top-5   accuracy (\%) & 83.39         & 5.23                                                                                                  & 80.50                                                                                                 & 57.38                                                                                                       \\ \bottomrule
    \end{tabular}
    \end{table}

\FloatBarrier


\end{document}